\documentclass[preprint,12pt,numbers,nopreprintline]{elsarticle}
\biboptions{sort&compress}

\usepackage{amsmath,amsfonts,amssymb}
\usepackage{mathtools}
\usepackage{algorithmic}
\usepackage{algorithm}
\usepackage{array}
\usepackage{textcomp}
\usepackage{url}
\usepackage{verbatim}
\usepackage{graphicx}
\graphicspath{{./}}

\usepackage[table]{xcolor}
\usepackage{float}
\usepackage{bm}
\usepackage{booktabs}
\usepackage{multirow}
\usepackage{threeparttable}
\usepackage{makecell}
\usepackage{rotating}
\usepackage{longtable}
\usepackage{pifont}
\usepackage[most]{tcolorbox}
\tcbuselibrary{raster,skins}
\usepackage[T1]{fontenc}
\usepackage[colorlinks=true,linkcolor=blue,citecolor=blue,urlcolor=blue]{hyperref}
\newcommand{\cmark}{\textcolor{green!60!black}{\ding{51}}}
\newcommand{\xmark}{\textcolor{red}{\ding{55}}}

\DeclareMathAlphabet\mathbfcal{OMS}{cmsy}{b}{n}

\DeclareMathAlphabet{\pazocal}{OMS}{zplm}{m}{n}
\DeclareMathAlphabet{\mathpzc}{OT1}{pzc}{m}{it}

\begin{document}

\begin{frontmatter}

\title{AgriScope: Pixel-Grounded Multimodal Understanding for Agricultural Images}

\author[cs]{Abderrahmene Boudiaf\corref{cor1}}
\ead{100058322@ku.ac.ae}
\author[cs]{Mohamad Alanssari}
\author[mne]{Irfan Hussain}
\author[cs]{Sajid Javed}
\cortext[cor1]{Corresponding author}

\affiliation[cs]{organization={Department of Computer Science, Khalifa University of Science and Technology},
                 addressline={P.O. Box 127788},
                 city={Abu Dhabi},
                 country={United Arab Emirates}}
\affiliation[mne]{organization={Department of Mechanical and Nuclear Engineering, Khalifa University of Science and Technology},
                  addressline={P.O. Box 127788},
                  city={Abu Dhabi},
                  country={United Arab Emirates}}

\begin{abstract}
Agricultural image understanding requires fine-grained recognition of plant diseases, pests, crop structures, and botanical species under complex real-world conditions.
Despite recent advances in Multimodal Large Language Models (MLLMs), existing models remain limited to text-only outputs and lack pixel-level visual grounding capabilities.
In this work, we introduce \textbf{AgriScope}, a unified pixel-grounded multimodal framework for agricultural image understanding.
AgriScope jointly supports image-level, region-level, and pixel-level understanding within a unified framework, enabling tasks such as grounded caption generation, referring expression segmentation, and multi-turn multimodal interaction for agricultural imagery.
AgriScope integrates biologically specialized semantic representations with dense spatial grounding through biological-semantic encoding, dense spatial representations, and pixel decoding.
To support large-scale grounded learning, we introduce \textbf{AgriGround}, a large-scale pixel-grounded agricultural multimodal instruction-tuning dataset containing over 500K images and 11M instruction-following samples spanning plant disease analysis, crop and weed identification, insect pest recognition, and fine-grained botanical understanding.
AgriGround is constructed through a multi-stage automatic annotation pipeline that integrates multimodal caption generation, phrase-level grounding, segmentation mask generation, and task-oriented instruction synthesis to produce densely grounded supervision.
Extensive experiments across multiple agricultural vision-language tasks demonstrate the effectiveness of AgriScope in pixel-grounded multimodal understanding, establishing a strong benchmark for agricultural vision-language learning and visual grounding.
The dataset and code will be made publicly available at \url{https://github.com/boudiafA/AgriScope}.
\end{abstract}


\end{frontmatter}

\section{Introduction}
\label{sec:introduction}
Agriculture plays a fundamental role in global food security \citep{godfray2010}, economic sustainability \citep{tilman2011}, and environmental stability \citep{tilman2011}.
Rapid advances in computer vision and AI have transformed modern agricultural systems by enabling automated crop monitoring \citep{agrivision}, plant disease diagnosis \citep{plantvillage}, precision farming \citep{agrivision}, yield estimation \citep{kamilaris2018}, pest analysis \citep{ip102}, weed management \citep{deepweeds}, and large-scale plant phenotyping \citep{inat2021}.
Recent developments in deep learning have significantly improved agricultural image analysis across diverse modalities, including field imagery \citep{plantdoc}, greenhouse monitoring \citep{kamilaris2018}, aerial sensing \citep{agrivision}, microscopy \citep{kamilaris2018}, and mobile-based crop diagnostics \citep{plantvillage}.
These models are increasingly being adopted in smart farming pipelines to reduce labor costs, improve productivity, and support data-driven agricultural decision-making \citep{kamilaris2018}.

Computer vision has emerged as an important domain for next-generation precision agriculture \citep{agrivision}.
Modern agricultural vision systems support a wide range of applications, including crop disease diagnosis \citep{plantvillage}, plant species identification \citep{inat2021}, fruit counting \citep{kamilaris2018}, crop and weed segmentation \citep{cropandweed}, pest recognition \citep{ip102}, irrigation monitoring \citep{kamilaris2018}, nutrient deficiency analysis \citep{kamilaris2018}, and robotic harvesting \citep{kamilaris2018}.
In practical agricultural deployments, visual understanding systems are used by farmers, agronomists, agricultural scientists, robotic platforms, and autonomous monitoring systems for large-scale crop analysis and decision support \citep{agrivision}.
Recent advances in deep learning and Vision-Language Models (VLMs) have further accelerated the development of intelligent agricultural systems capable of understanding both visual and textual agricultural information. 
Large-scale vision transformers, multimodal representation learning, and Multimodal Large Language Models (MLLMs) have demonstrated promising capabilities for agricultural image captioning \citep{agrogpt}, Visual Question Answering (VQA) \citep{agrogpt}, and disease classification \citep{plantvillage}.
Existing models provide a new paradigm for interactive agricultural intelligence by enabling natural-language interaction with agricultural imagery \citep{agrogpt}.

\begin{figure*}[!htbp]
    \centering
    \includegraphics[width=\textwidth]{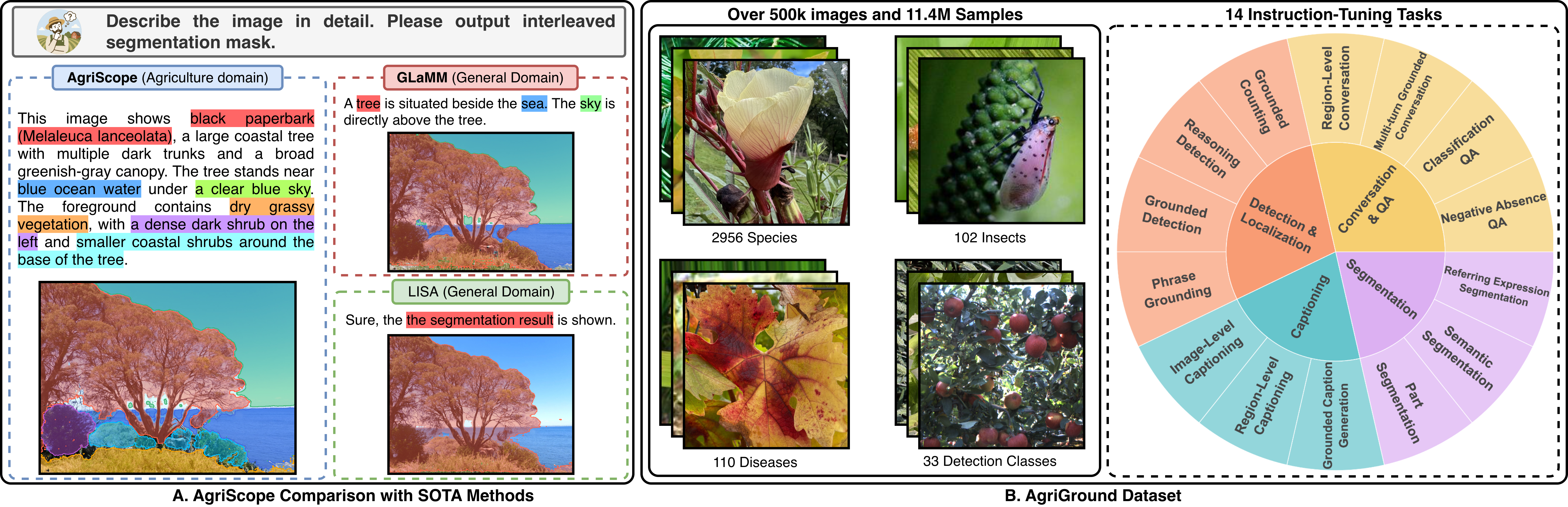}
    \caption{Overview of AgriScope and the AgriGround dataset. (a) Comparison of the proposed AgriScope model with state-of-the-art MLLMs. (b) Summary of the proposed AgriGround dataset and its diverse multimodal tasks.}
    \label{fig:AgriGround_overview}
\end{figure*}

\begin{table*}[t]
  \centering
  \caption{Comparison of the proposed AgriGround dataset with representative agricultural vision-only and vision-language datasets in terms of tasks and domain coverage.}
  \label{tab:dataset_comparison}
  \resizebox{\textwidth}{!}{%
  \begin{threeparttable}
  \begin{tabular}{@{}llrr ccccccc cccc@{}}
    \toprule
    \multirow{2}{*}{\textbf{Dataset}} &
    \multirow{2}{*}{\textbf{Venue}} &
    \multirow{2}{*}{\textbf{\#Images}} &
    \multirow{2}{*}{\textbf{\#Samples}} &
    \multicolumn{7}{c}{\textbf{Tasks}} &
    \multicolumn{4}{c}{\textbf{Domain Coverage}} \\
    \cmidrule(lr){5-11}\cmidrule(lr){12-15}
    & & & &
    \textbf{VQA} & \textbf{Cap.} & \textbf{Dial.} &
    \textbf{GCG} & \textbf{RES} & \textbf{PG} & \textbf{Count.} &
    \textbf{Disease} & \textbf{Weed} & \textbf{Insect} & \textbf{Species} \\
    \midrule

    \multicolumn{15}{l}{\textit{Vision-Only Datasets}} \\[2pt]

    PlantVillage~\citep{plantvillage}
      & arXiv 2015     &  54{,}306              &  54{,}306
      & \xmark & \xmark & \xmark & \xmark & \xmark & \xmark & \xmark
      & \cmark & \xmark & \xmark & \xmark \\

    PlantDoc~\citep{plantdoc}
      & CODS 2020      &   2{,}598              &   8{,}595
      & \xmark & \xmark & \xmark & \xmark & \xmark & \xmark & \xmark
      & \cmark & \xmark & \xmark & \xmark \\

    DeepWeeds~\citep{deepweeds}
      & Sci.Rep.\ 2019 &  17{,}509              &  17{,}509
      & \xmark & \xmark & \xmark & \xmark & \xmark & \xmark & \xmark
      & \xmark & \cmark & \xmark & \xmark \\

    CropAndWeed~\citep{cropandweed}
      & WACV 2023      &   8{,}000              & 112{,}000
      & \xmark & \xmark & \xmark & \xmark & \xmark & \xmark & \xmark
      & \xmark & \cmark & \xmark & \xmark \\

    IP102~\citep{ip102}
      & CVPR 2019      &  75{,}000              &  75{,}000
      & \xmark & \xmark & \xmark & \xmark & \xmark & \xmark & \xmark
      & \xmark & \xmark & \cmark & \xmark \\

    Agriculture-Vision~\citep{agrivision}
      & CVPR 2020      &  94{,}986              &  94{,}986
      & \xmark & \xmark & \xmark & \xmark & \xmark & \xmark & \xmark
      & \cmark & \xmark & \xmark & \xmark \\

    iNaturalist-2021~\citep{inat2021}
      & FGVC 2021      & \textbf{2{,}700{,}000} & 2{,}700{,}000
      & \xmark & \xmark & \xmark & \xmark & \xmark & \xmark & \xmark
      & \xmark & \xmark & \xmark & \cmark \\

    \midrule

    \multicolumn{15}{l}{\textit{Vision-Language Datasets}} \\[2pt]

    AgroInstruct~\citep{agrogpt}
      & WACV 2025      & $\sim$6{,}000          &  70{,}000
      & \cmark & \cmark & \cmark & \xmark & \xmark & \xmark & \xmark
      & \cmark & \cmark & \cmark & \xmark \\

    AgriCLIP / ALive~\citep{agriclip}
      & COLING 2025    & 600{,}000              & 600{,}000
      & \xmark & \cmark & \xmark & \xmark & \xmark & \xmark & \xmark
      & \cmark & \xmark & \xmark & \xmark \\

    CDDM~\citep{cddm}
      & ECCV 2024      & 137{,}000              & 1{,}000{,}000
      & \cmark & \xmark & \cmark & \xmark & \xmark & \xmark & \xmark
      & \cmark & \xmark & \xmark & \xmark \\

    Agri-LLaVA~\citep{agrillava}
      & arXiv 2024     & 400{,}000              & 400{,}000
      & \cmark & \cmark & \cmark & \xmark & \xmark & \xmark & \xmark
      & \cmark & \xmark & \cmark & \xmark \\

    AgriDoctor / AgriMM~\citep{agridoctor}
      & arXiv 2025     & 400{,}000              & 300{,}000
      & \cmark & \cmark & \cmark & \xmark & \xmark & \xmark & \xmark
      & \cmark & \xmark & \xmark & \xmark \\

    Agri-3M-VL~\citep{agri3mvl}
      & arXiv 2025     & $\sim$500{,}000\tnote{$\dagger$} & 3{,}065{,}000
      & \cmark & \cmark & \xmark & \xmark & \xmark & \xmark & \xmark
      & \cmark & \cmark & \cmark & \xmark \\

    AgroBench~\citep{agrobench}
      & ICCV 2025      &   3{,}745              &   4{,}342
      & \cmark & \xmark & \xmark & \xmark & \xmark & \xmark & \xmark
      & \cmark & \cmark & \cmark & \xmark \\

    AgMMU~\citep{agmmu}
      & NeurIPS 2025   & 1{,}094\tnote{$\ddagger$} &  57{,}079\tnote{$\S$}
      & \cmark & \xmark & \xmark & \xmark & \xmark & \xmark & \xmark
      & \cmark & \xmark & \cmark & \cmark \\

    AgroMind~\citep{agromind}
      & arXiv 2025     &  20{,}850              &  28{,}482
      & \cmark & \xmark & \xmark & \xmark & \xmark & \xmark & \xmark
      & \cmark & \cmark & \xmark & \xmark \\

    \midrule

    \rowcolor[gray]{0.92}
    \textbf{AgriGround (Ours)}
      & \textemdash    & 503{,}919              & \textbf{11{,}421{,}148}
      & \cmark & \cmark & \cmark & \cmark & \cmark & \cmark & \cmark
      & \cmark & \cmark & \cmark & \cmark \\
    \bottomrule
  \end{tabular}
  \begin{tablenotes}
   \small
   \item[] \textit{Abbreviations and symbols:}
      VQA = visual question answering, Cap. = captioning, Dial. = dialogue,
      GCG = grounded caption generation, RES = referring expression segmentation,
      PG = phrase grounding, Count. = counting, MCQ = multiple-choice question.
    \item[$\dagger$] Agri-3M-VL aggregates images from nine public and one private source;
     the exact unique image count is not publicly reported.
    \item[$\ddagger$] AgMMU reports 1,094 images in its public MCQ evaluation set;
     the full corpus is derived from 116,231 real-world multimodal dialogues,
     for which a flat image total is not officially reported.
    \item[$\S$] AgMMU \#Samples corresponds to the 57,079 verified multimodal
     facts in AgBase, the accompanying training knowledge base.
          The MCQ benchmark itself contains 746 questions (plus 746 open-ended equivalents).
    \item[] \textbf{Highest value} in each numeric column is shown in bold.
  \end{tablenotes}
  \end{threeparttable}}
\end{table*}

Early agricultural computer vision systems primarily focused on supervised image classification and object detection using CNNs \citep{he2016resnet}.
The available datasets including PlantVillage \citep{plantvillage}, PlantDoc \citep{plantdoc}, DeepWeeds \citep{deepweeds}, IP102 \citep{ip102}, and Agriculture-Vision \citep{agrivision} facilitated research in disease recognition \citep{plantvillage}, weed detection \citep{deepweeds}, pest analysis \citep{ip102}, and aerial agricultural monitoring \citep{agrivision}.
More recently, foundation models have further improved generalization and scalability in agricultural image understanding \citep{bioclip2,sam3}.
The emergence of VLMs and MLLMs has further expanded agricultural imagery research \citep{agrogpt,agrillava}.
Recent agricultural VLMs and MLLMs, including AgroGPT \citep{agrogpt}, Agri-LLaVA \citep{agrillava}, AgriGPT-VL \citep{agri3mvl}, and AgriDoctor \citep{agridoctor}, integrate visual-textual understanding for tasks such as VQA and agricultural dialogue. Their associated instruction-tuning corpora, including AgroInstruct \citep{agrogpt} and Agri-3M-VL \citep{agri3mvl}, support the development of domain-specialized models (Fig.~\ref{fig:AgriGround_overview}).
In parallel, generic visual grounding models such as LISA \citep{lisa}, GLaMM \citep{glamm}, and Osprey \citep{osprey} have demonstrated impressive capabilities in visual grounding and segmentation-aware multimodal interaction.
 
Although existing MLLMs have shown promising results, they remain limited in several important aspects. 
\textit{First}, state-of-the-art agricultural MLLMs operate purely in text space and do not spatially ground semantic concepts within agricultural images.
Current models describe diseases, pests, or crop conditions in natural language, but do not localize or segment the corresponding regions at the pixel level.
\textit{Second}, existing agricultural datasets lack dense grounding annotations required for fine-grained multimodal understanding \citep{plantvillage,ip102,agrivision}. 
Most available datasets are restricted to image-level labels, object detection annotations, or textual descriptions, without aligned segmentation masks, grounded phrases, and multimodal instruction supervision \citep{plantvillage,deepweeds,ip102,agrivision}.
Consequently, existing models fail to support grounded agricultural tasks such as referring expression segmentation \citep{lisa}, phrase grounding \citep{glamm}, grounded caption generation \citep{glamm}, grounded counting \citep{glamm}, or pixel-level multimodal interaction \citep{lisa}.
\textit{Third}, agricultural imagery introduces domain-specific challenges that are not adequately addressed by generic grounding models.
Fine-grained plant pathology, crop morphology, insect appearance, and lesion-level analysis require biologically informed visual representations and high-resolution spatial grounding mechanisms beyond generic natural-image understanding.
\textit{These limitations reveal a significant research gap in agricultural MLLMs: the absence of large-scale pixel-grounded agricultural datasets and unified multimodal frameworks capable of jointly supporting image-level, region-level, and pixel-level agricultural understanding.}

Visual grounding is particularly important for agricultural applications because agricultural decision-making often depends on the precise localization of visual symptoms and structures \citep{glamm}.
Farmers and agronomists require not only textual descriptions but also accurate localization of infected plant regions, disease lesions, insect infestations, weed boundaries, damaged organs, and crop structures \citep{cropandweed,ip102}. 
Pixel-grounded MLLMs may significantly improve interpretability, transparency, and reliability in agricultural AI pipelines by explicitly linking semantic concepts to visual evidence.
Grounded agricultural understanding also enables a new generation of interactive agricultural assistants capable of answering spatially aware questions such as ``\textit{Where is the disease located?}'', ``\textit{Can you segment the infected region?}'', ``\textit{Which leaves contain pest damage?}'', and ``\textit{How many wheat heads are visible in this image?}'' Such capabilities are essential for precision agriculture, automated crop monitoring, robotic farming, agricultural robotics, and explainable agricultural decision-support systems \citep{agrivision,kamilaris2018}.
 
Motivated by these challenges, we present a unified pixel-grounded MLLM and a large-scale grounded agricultural dataset that jointly advance agricultural vision-language learning and visual grounding.
In this work, we introduce \textbf{AgriScope}, a unified pixel-grounded MLLM  for agricultural image understanding and analysis. 
AgriScope jointly supports image-level, region-level, and pixel-level understanding within a single end-to-end architecture. 
Unlike existing agricultural MLLMs that produce only global predictions, such as textual outputs, AgriScope generates pixel-grounded outputs directly conditioned on natural-language prompts.
AgriScope supports a broad range of agricultural tasks, ranging from grounded caption generation to multi-turn multimodal interaction. 
The framework combines biologically specialized semantic representations with dense spatial grounding through BioCLIP~2-based biological-semantic encoding, DINOv3-based dense spatial representations, and SAM~2-based pixel decoding within a unified MLLM.
For pixel-level grounding, AgriScope employs a SAM~2-based segmentation decoder conditioned on segmentation tokens generated by the MLLM.
This design enables direct generation of segmentation masks conditioned on open-vocabulary natural-language prompts. 
This unified architecture allows AgriScope to support these diverse tasks within a single framework.

To support large-scale grounded agricultural learning, we further introduce \textbf{AgriGround}, the largest automatically annotated pixel-grounded agricultural multimodal instruction-tuning dataset to date. AgriGround contains over 500K agricultural images and more than 11M instruction-following samples spanning plant disease analysis, crop and weed identification, insect pest recognition, plant-part segmentation, and fine-grained botanical understanding (Table~\ref{tab:dataset_comparison}).
The dataset is constructed through a multi-stage automatic annotation pipeline that integrates multimodal caption generation, phrase-level grounding, segmentation mask generation, and task-oriented instruction synthesis. 
Our annotation pipeline produces densely grounded multimodal supervision in the form of captions, grounded object phrases, segmentation masks, and bounding boxes. 
AgriGround supports a diverse set of grounded multimodal tasks, including referring expression segmentation, phrase grounding, grounded detection, semantic segmentation, and grounded counting.
 
We evaluate AgriScope on AgriGround across multiple agricultural vision-language and visual grounding tasks, as well as on other publicly available datasets.
Extensive experiments demonstrate that AgriScope consistently outperforms existing agricultural and general-purpose MLLMs on agricultural image understanding tasks.
Both quantitative and qualitative evaluations show the effectiveness of AgriScope in fine-grained pixel grounding, multimodal alignment, and agricultural visual understanding.
Despite significant progress, agricultural image understanding remains substantially more challenging than generic natural-image understanding \citep{plantdoc}.
Agricultural imagery exhibits extreme intra-class variability caused by environmental conditions, illumination changes, occlusions, seasonal variations, disease progression stages, and complex field backgrounds \citep{plantdoc}.
Fine-grained differences between plant species, diseases, pests, and crop structures further increase the difficulty of robust visual understanding \citep{ip102}.
Agricultural datasets are often long-tailed, highly imbalanced, and domain-specific (Table~\ref{tab:dataset_comparison}) \citep{ip102}.
Disease symptoms may appear as extremely small lesions occupying only a tiny fraction of the image, while field scenes frequently contain overlapping leaves, cluttered vegetation, and dense object distributions \citep{plantdoc}.
Moreover, many agricultural applications require fine-grained spatial localization rather than global image classification alone \citep{glamm}.
For example, agronomists often need to identify the exact infected region on a leaf, localize pest infestations, segment plant organs, or distinguish crops from weeds at the pixel level \citep{glamm}.
These challenges make agricultural visual understanding fundamentally different from generic image recognition tasks.

The main contributions of our work are summarized as follows:

\begin{itemize}
    \item We introduce AgriScope, a unified pixel-grounded MLLM for agricultural image understanding supporting image-level, region-level, and pixel-level multimodal interaction within a single architecture.
    \item We present AgriGround, the largest automatically annotated pixel-grounded agricultural multimodal instruction-tuning dataset comprising over 500K images and 11M instruction-following samples with dense grounding supervision.
   \item We propose a multi-stage automatic annotation pipeline that integrates multimodal caption generation, phrase grounding, segmentation mask generation, and instruction synthesis for scalable agricultural multimodal supervision.
   \item Extensive experiments demonstrate the effectiveness of AgriScope across multiple agricultural multimodal understanding and visual grounding tasks, establishing a strong benchmark for future agricultural vision-language learning research.
\end{itemize}

The remainder of this paper is organized as follows. 
Section~\ref{sec:related} reviews related work in agricultural image analysis.
Section~\ref{sec:method} presents the proposed AgriScope MLLM.
Section~\ref{sec:dataset} describes the AgriGround dataset, the proposed
automatic annotation pipeline, and the experimental evaluation, while
Section~\ref{sec:conclusion} concludes the paper and discusses future research
directions.

\begin{table*}[t]
  \centering
  \caption{Comparison of AgriScope with existing agricultural MLLMs.}
  \label{tab:model_comparison_ag}
  \resizebox{\textwidth}{!}{%
  \begin{threeparttable}
  \begin{tabular}{@{}
      l        
      l        
      l        
      l        
      >{\centering\arraybackslash}p{1.8cm}   
      >{\centering\arraybackslash}p{2.2cm}   
      >{\centering\arraybackslash}p{2.2cm}   
      c c c c c c c c c                      
      @{}}
    \toprule
    \multirow{2}{*}{\textbf{Model}} &
    \multirow{2}{*}{\textbf{Venue}} &
    \multirow{2}{*}{\textbf{Base LLM}} &
    \multirow{2}{*}{\textbf{Global Enc.}} &
    \multirow{2}{*}{\textbf{Bio. Enc.}} &
    \multirow{2}{*}{\textbf{\makecell{BBox\\Grounded\\Output}}} &
    \multirow{2}{*}{\textbf{\makecell{Pixel-Level\\Grounded\\Output}}} &
    \multicolumn{9}{c}{\textbf{Supported Tasks}} \\
    \cmidrule(l){8-16}
    & & & & & & &
    \rotatebox{75}{\textbf{Cls}} &
    \rotatebox{75}{\textbf{Pest}} &
    \rotatebox{75}{\textbf{Bot}} &
    \rotatebox{75}{\textbf{VQA}} &
    \rotatebox{75}{\textbf{Cap}} &
    \rotatebox{75}{\textbf{RES}} &
    \rotatebox{75}{\textbf{PSeg}} &
    \rotatebox{75}{\textbf{SSeg}} &
    \rotatebox{75}{\textbf{Cnt}} \\
    \midrule

    AgroGPT~\citep{agrogpt}
      & WACV 2025 & LLaMA-3-8B & CLIP ViT-L/14
      & \textemdash & \xmark & \xmark
      & \cmark & \cmark & \cmark & \cmark & \xmark & \xmark & \xmark & \xmark & \xmark \\

    Agri-LLaVA~\citep{agrillava}
      & arXiv 2024 & LLaMA-3-8B & CLIP ViT-L/14
      & \textemdash & \xmark & \xmark
      & \cmark & \cmark & \xmark & \cmark & \xmark & \xmark & \xmark & \xmark & \xmark \\

    CDDM~\citep{cddm}
      & ECCV 2024 & Qwen-VL & ViT-bigG (OpenCLIP)
      & \textemdash & \xmark & \xmark
      & \cmark & \xmark & \xmark & \cmark & \xmark & \xmark & \xmark & \xmark & \xmark \\

    AgriGPT-VL~\citep{agri3mvl}
      & arXiv 2025 & Qwen2.5-VL-7B & SigLIP-SO400M
      & \textemdash & \xmark & \xmark
      & \cmark & \xmark & \cmark & \cmark & \xmark & \xmark & \xmark & \xmark & \xmark \\

    AgriChat~\citep{agrichat}
      & arXiv 2026 & Qwen2.5-VL & SigLIP-SO400M
      & \textemdash & \xmark & \xmark
      & \cmark & \cmark & \cmark & \cmark & \xmark & \xmark & \xmark & \xmark & \xmark \\

    \midrule
    \rowcolor{gray!12}
    AgriScope (Ours)
      & \textemdash & Qwen2.5-0.5B-Instruct & DINOv3
      & BioCLIP~2 & \cmark & \cmark
      & \cmark & \cmark & \cmark & \cmark & \cmark & \cmark & \cmark & \cmark & \cmark \\
    \bottomrule
  \end{tabular}

  \begin{tablenotes}[flushleft]
    \footnotesize
    \item[] \textit{Abbreviations and symbols:}
    Enc. = encoder,
    BBox = bounding box,
    Cls = classification,
    Pest = pest recognition,
    Bot = botanical species recognition,
    VQA = visual question answering,
    Cap = captioning,
    RES = referring expression segmentation,
    PSeg = part segmentation,
    SSeg = semantic segmentation, and
    Cnt = counting.
  \end{tablenotes}
  \end{threeparttable}%
  }
\end{table*}

\section{Related Work}
\label{sec:related}
Recent years have witnessed significant progress in agricultural computer vision, resulting in a wide range of specialized datasets and learning frameworks tailored to agricultural applications \citep{kamilaris2018, agrivision, agrobench}. 
Below, we discuss representative agricultural vision models, Vision-Language Models (VLMs), Multimodal Large Language Models (MLLMs), agricultural datasets, and the key tasks addressed in the literature.
 
\subsection{Agricultural Vision Models}
\label{sec:related:vision}
Computer vision for agriculture has been transformed over the past decade by leveraging deep learning models for image-level tasks \citep{kamilaris2018}. 
Plant disease classification models based on VGG- and ResNet-driven architectures trained on the PlantVillage dataset \citep{plantvillage} achieved 99.35\% in-distribution accuracy on cropped, lab-style leaf imagery \citep{plantvillage}, yet their performance collapsed below 40\% on in-the-wild images.
Cluttered canopy scenes drove parallel efforts in crop/weed discrimination, where single-stage and two-stage detectors, including YOLO \citep{redmon2016yolo} variants and Faster R-CNN backbones \citep{ren2015fasterrcnn}, have been adopted for plant localization \citep{deepweeds, agrivision}.
DeepWeeds contributed one of the earliest large-scale, in-situ, multiclass benchmarks for rangeland weed recognition \citep{deepweeds}, and the more recent CropAndWeed dataset \citep{cropandweed} extended this paradigm through fine-grained, multimodal supervision covering 74 crop and weed species annotated with bounding boxes and semantic masks~\citep{cropandweed}.
 
Self-attention mechanisms subsequently constituted a second wave of progress in agricultural computer vision \citep{swin2021}.
The Vision Transformer (ViT) architecture and its variants have been adapted to plant phenotyping \citep{dosovitskiy2021vit}, fine-grained disease classification \citep{plantdoc}, and pest recognition \citep{ip102}, improving performance on subtle, low-contrast lesions and long-range stem and leaf structures \citep{swin2021}.
Hybrid CNN-Transformer architectures, frequently paired with attention-based fusion modules, have since become standard across agricultural classification leaderboards \citep{swin2021}.
Segment Anything (SAM) \citep{sam} and domain-adapted SAM variants \citep{sam2} have also enabled crop-head counting, field-boundary delineation from Sentinel-2 imagery, and weed and disease mask annotation, while DINO-based self-supervised learning (SSL) has further improved performance across detection, segmentation, and classification tasks on a variety of agricultural imagery datasets \citep{dinoV3, bioclip, bioclip2}.
 
Existing state-of-the-art models have improved performance; however, three structural limitations remain \citep{kamilaris2018, agrobench}.
First, a disease classifier may not count tillers, a weed detector may not describe symptoms, and a segmentation backbone may not answer
``why'' questions because these systems are designed for task-specific prediction \citep{deepweeds, agrivision, cropandweed}.
Second, existing agricultural models do not leverage natural-language assistance tightly coupled to perception tasks \citep{deepweeds, agrivision, cropandweed}.
Third, pixel-level reasoning conditioned on language remains absent in agricultural tasks \citep{agrogpt, agrillava}.
\textit{In this work, we advance the field by presenting a visual-grounding MLLM for agricultural imagery and a pixel-grounded dataset that supports a wide range of agricultural tasks.}
 
 \begin{figure*}[t!]
  \centering
  \includegraphics[width=0.95\linewidth]{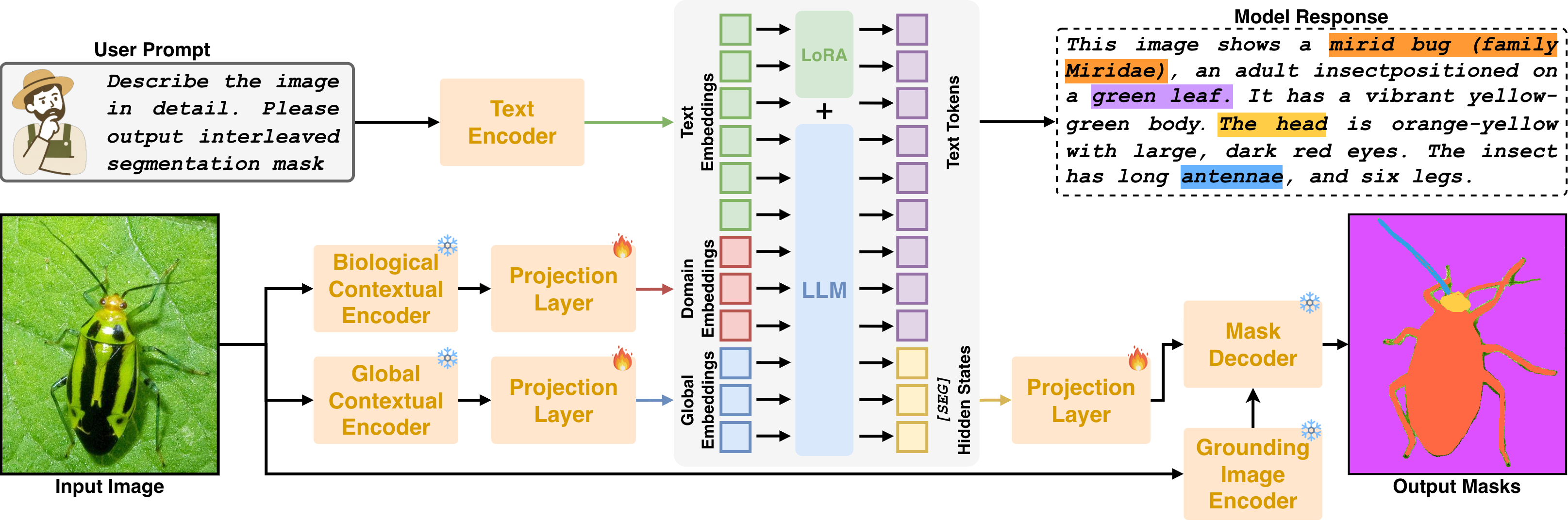}
  \caption{\textbf{Overview of the proposed AgriScope architecture.} Given an agricultural image and a natural-language prompt, AgriScope extracts complementary visual representations using a BioCLIP~2-based biological semantic encoder and a DINOv3-based global contextual encoder. The resulting visual embeddings are projected into a shared multimodal embedding space and fused with textual embeddings before being processed by a LoRA-adapted large language model (LLM). The LLM generates both textual responses and special \texttt{[SEG]} tokens corresponding to grounded entities. The hidden representations of the \texttt{[SEG]} tokens are projected to condition a SAM~2-driven mask decoder, which produces pixel-level segmentation masks aligned with the generated response. Snowflake and flame icons denote frozen and trainable modules, respectively.}
  \label{fig:agriscope_architecture}
\end{figure*}

\subsection{VLMs and MLLMs for Agricultural Imagery}
\label{sec:related:vlm}
Inspired by the success of foundation VLMs, recent research has increasingly explored domain-specific adaptation of VLMs and MLLMs for agricultural applications \citep{agriclip, agrogpt, agrillava, agridoctor, agrichat}. 
These models aim to bridge visual perception and agricultural knowledge, enabling natural-language interaction with crop imagery, disease symptoms, pest infestations, and field scenes \citep{agrogpt, agrillava, agridoctor, agrichat}. Compared to traditional agricultural computer vision systems that focus on single tasks such as disease classification or object detection, agricultural VLMs provide a unified framework for multimodal understanding and human-AI interaction.
 
Several studies have adapted large-scale vision-language pretraining to agricultural domains. 
AgriCLIP \citep{agriclip} and related domain-specific representation learning approaches improve cross-modal alignment by leveraging agricultural image-text pairs, enabling robust crop and disease recognition under varying environmental conditions.
Building upon these representations, recent MLLMs, including AgroGPT \citep{agrogpt}, Agri-LLaVA \citep{agrillava}, Agri-Doctor \citep{agridoctor}, and AgriChat \citep{agrichat}, extend agricultural understanding to visual question answering, image captioning, multimodal dialogue, knowledge-assisted diagnosis, and instruction-following tasks. 
These models demonstrate the potential of MLLMs as intelligent agricultural assistants capable of providing descriptive and diagnostic information from agricultural imagery.
A brief summary of state-of-the-art agriculture-specific MLLMs is provided in Table~\ref{tab:model_comparison_ag}.
 
Despite these advances, existing agricultural VLMs and MLLMs remain primarily text-centric \citep{agrogpt, agrillava, agridoctor, agrichat}. 
Their outputs are typically limited to image-level descriptions, question answering, or conversational responses, without explicit spatial grounding of agricultural concepts.
Consequently, existing models may not identify the precise location of disease symptoms, localize pest infestations, segment plant organs, distinguish crops from weeds at the pixel level, or associate textual descriptions with visual evidence \citep{agrogpt, agrillava, agridoctor, agrichat}.
\textit{Such limitations significantly reduce interpretability and practical utility in precision agriculture, where accurate localization is often as important as semantic recognition.}
These limitations highlight a critical research gap in agricultural imagery, including the lack of a unified grounding-capable MLLM. 
\textit{This work addresses the gap by introducing AgriScope, a unified MLLM framework that enables image-level, region-level, and pixel-level agricultural understanding through visual grounding.}

\subsection{Agricultural Imagery Datasets and Tasks}
\label{sec:related:data}
Existing agricultural datasets are primarily designed for individual computer vision or vision-language tasks, providing image-level labels for classification \citep{plantvillage, plantdoc, ip102}, bounding-box annotations for object detection \citep{deepweeds, agrivision, cropandweed}, pixel-level masks for segmentation \citep{cropandweed, agrivision}, or image-text pairs for multimodal learning \citep{agrogpt, cddm, agrillava} (Table~\ref{tab:dataset_comparison}).
While these datasets have significantly advanced agricultural image understanding, they lack dense visual grounding annotations that explicitly associate semantic concepts with their spatial locations in an image \citep{glamm, lisa}.
To address this limitation, we introduce \textbf{AgriGround}, a large-scale pixel-grounded multimodal instruction-tuning dataset for agricultural image understanding. 
AgriGround contains more than 500K agricultural images and over 11M instruction-following samples spanning diverse agricultural domains.
Unlike existing agricultural datasets, AgriGround jointly provides natural-language descriptions, grounded object phrases, segmentation masks, and spatial annotations, enabling direct alignment between visual regions and semantic concepts.
AgriGround supports a diverse set of image-level, region-level, and pixel-level tasks under a unified multimodal learning framework enabling comprehensive evaluation of agricultural multimodal systems.
Compared with existing agricultural vision-language datasets such as AgroInstruct \citep{agrogpt}, Agri-LLaVA \citep{agrillava}, AgroMind \citep{agromind}, and AgriDoctor \citep{agridoctor}, AgriGround is the first dataset to provide large-scale pixel-grounded supervision for agricultural multimodal learning.

\section{Methodology}
\label{sec:method}
\subsection{Overview of AgriScope Model}
We introduce \textbf{AgriScope}, a unified pixel-grounded multimodal large language model (MLLM) for agricultural image understanding.
Unlike existing agricultural MLLMs that generate only image-level textual responses, AgriScope establishes explicit correspondences between natural-language concepts and their visual regions. It therefore recognizes agricultural entities such as diseases, pests, and weeds while accurately localizing and segmenting them at the image, region, and pixel levels.
A system diagram of our proposed AgriScope model is illustrated in Fig.~\ref{fig:agriscope_architecture}.
AgriScope adopts a multi-component architecture consisting of a biological contextual visual encoder, a global contextual visual encoder, a large language model (LLM), and a pixel-grounding decoder.
The biological contextual visual encoder captures high-level semantic information related to plant species and disease symptoms, while the global contextual visual encoder extracts fine-grained spatial representations.
These visual features are integrated with textual features through an LLM that performs multimodal reasoning and instruction following. 
For pixel-level localization, AgriScope employs a SAM~2-driven decoder conditioned on language representations, enabling direct generation of segmentation masks from natural-language prompts.
AgriScope is pre-trained in two stages on the proposed AgriGround dataset, including visual-language alignment with grounding-aware representation learning and multimodal instruction tuning using diverse grounded tasks. 
AgriScope provides a comprehensive framework for next-generation agricultural vision-language understanding and precision agriculture applications.

\subsection{Problem Formulation}
Let $I \in \mathbb{R}^{H \times W \times 3}$ denote an agricultural image and let $T=\{t_1,t_2,\ldots,t_n \}$ denote a natural-language instruction describing a user query. The objective of AgriScope is to learn a unified multimodal mapping
$\mathcal{F} : (I,T) \rightarrow (Y,\mathcal{M},\mathcal{B})$, where $Y$ denotes the generated textual response, $\mathcal{M}=\{ M_1,\ldots,M_K \}$ denotes a set of pixel-level segmentation masks, and $\mathcal{B}=\{ B_1,\ldots,B_K \}$ denotes the corresponding spatial grounding information.
Unlike state-of-the-art agricultural VLMs that generate only textual outputs, AgriScope explicitly associates semantic concepts appearing in the response with their corresponding visual regions.
Let $\mathcal{C}=\{c_1,c_2,\ldots,c_K\}$ denote the set of grounded concepts referenced in the generated response, where each concept may correspond to an agricultural entity such as a disease symptom, pest, crop, weed, plant organ, or botanical species. AgriScope establishes a mapping $c_k \rightarrow (M_k,B_k)$, which grounds each semantic concept $c_k$ to its corresponding image region through a segmentation mask $M_k$ and optional bounding box $B_k$.
This formulation supports multiple levels of agricultural understanding.
For instance, at the image-level, the model generates a holistic description or answer, $\mathcal{F}_{img}:(I,T)\rightarrow Y$.
At the region-level, the model conditions on a user-specified region $R$, $\mathcal{F}_{reg}:(I,R,T)\rightarrow Y_R$, where $Y_R$ denotes a region-specific response.
At the pixel-level, the model performs visual grounding, $\mathcal{F}_{pix}:(I,T)\rightarrow (\mathcal{M},\mathcal{B})$, enabling localization and segmentation of agricultural concepts referenced in natural-language expressions.
The goal of AgriScope is therefore to unify image-level, region-level, and pixel-level agricultural understanding within a single multimodal framework capable of generating both semantic and spatially grounded outputs.

\subsection{AgriScope Main Architecture}
AgriScope consists of four main components: (i) a biological contextual encoder, (ii) a global visual encoder, (iii) a multimodal large language model (LLM), and (iv) a grounded segmentation decoder.

\subsubsection{\textbf{Biological Contextual Visual Encoder}}
To capture domain-specific visual representations, AgriScope employs BioCLIP~2 \citep{bioclip2} as a biological semantic encoder that extracts high-level contextual representations from agricultural images.
Given an input image $I$, the encoder generates a set of biological contextual visual tokens as:

\begin{equation}
\mathbf{V}_{bio} = E_{bio}(I), 
\end{equation}
\noindent where $\mathbf{V}_{bio} \in \mathbb{R}^{N_b \times d}$ denotes biologically informed visual representations. 
This biological semantic representation captures agricultural concepts and fine-grained visual characteristics that are critical for downstream agricultural understanding tasks and serve as the primary source of semantic knowledge for multimodal reasoning.

\subsubsection{\textbf{Global Contextual Visual Encoder}}
While the biological semantic encoder captures high-level agricultural concepts and language-aligned semantics, accurate visual grounding additionally requires fine-grained spatial representations that preserve local image structures. 
To this end, AgriScope incorporates a DINOv3-based ViT visual encoder~\citep{dinoV3} to extract dense contextual features from the input image.
Given an agricultural image $I$, the encoder generates a set of visual embeddings as:

\begin{equation}
\mathbf{V}_{gl} = E_{gl}(I),
\end{equation}

\noindent where $\mathbf{V}_{gl} \in \mathbb{R}^{N_r \times d}$ denotes the dense visual representations associated with image regions.
Unlike the BioCLIP~2-based semantic encoder, which focuses on language-aligned agricultural semantics, the DINOv3 encoder preserves detailed visual characteristics such as texture, morphology, boundaries, and structural patterns.
These representations are particularly important for identifying disease lesions, plant organs, insect pests, crop instances, and other localized agricultural phenomena.
The complementary nature of the two visual streams enables AgriScope to jointly model \emph{what} is present in the image and \emph{where} it is located.
While the biological semantic encoder provides concept-level understanding, the DINOv3 encoder provides precise spatial information necessary for localization and pixel-level grounding. 
Consequently, the resulting visual representations maintain strong spatial correspondence between semantic concepts and their associated image regions, facilitating accurate visual grounding and segmentation.

\subsubsection{\textbf{Text Encoder}} 
AgriScope supports natural-language interaction through the text encoder of Qwen2.5-0.5B-Instruct \citep{qwen2024qwen25_05b_instruct,qwen2025qwen25technicalreport}.
Given an input instruction or query $T$, the text encoder converts the input text into a sequence of token embeddings as:

\begin{equation}
\mathbf{E}_{txt} = E_{txt}(T),    
\end{equation}

\noindent where $\mathbf{E}_{txt} \in \mathbb{R}^{N_t \times d}$ denotes the textual representations and $N_t$ is the number of input tokens. 
These embeddings provide a semantic representation of the user's instruction and serve as the textual input to the MLLM.

\subsubsection{\textbf{Large Language Model}}
The LLM serves as the central reasoning and instruction-following component of AgriScope.
We adopt Qwen2.5-0.5B-Instruct \citep{qwen2024qwen25_05b_instruct} as the language backbone because of its strong reasoning capabilities and efficient deployment characteristics.
The biological semantic tokens $\mathbf{V}_{bio}$, global dense tokens $\mathbf{V}_{gl}$, and the textual embeddings $\mathbf{E}_{txt}$ are first projected into a common embedding space and concatenated along the sequence dimension. 
The resulting multimodal token sequence is processed by the language model as:

\begin{equation}
\mathbf{H}=
\operatorname{LLM}
\left(
\mathbf{V}_{bio}
\mathbin{\Vert}
\mathbf{V}_{gl}
\mathbin{\Vert}
\mathbf{E}_{txt}
\right)
\in
\mathbb{R}^{(N_b + N_r + N_t)\times d_{\ell}},
\label{eq:llm}
\end{equation}

\noindent where $\Vert$ denotes sequence-level concatenation, $N_b$ and $N_r$ denote the numbers of biological-semantic and spatial visual tokens, respectively, $N_t$ is the number of textual tokens, and $d_{\ell}$ represents the hidden dimension of the language model.
The resulting hidden representations $\mathbf{H}$ capture interactions between agricultural visual concepts and natural-language instructions, enabling image captioning, VQA, grounded conversation, disease diagnosis, and multimodal agricultural understanding. 
The LLM autoregressively generates output tokens conditioned on both visual-textual inputs.

For visual grounding tasks, AgriScope introduces a special segmentation token \texttt{[SEG]} that indicates concepts requiring pixel-level localization.
Let $\mathbf{h}_{[\mathrm{SEG}]}^{(k)} \in \mathbb{R}^{d_\ell}$ denote the final hidden representation corresponding to the $k$-th \texttt{[SEG]} token. A learnable projection function

\begin{equation}
\mathbf{c}_k = \psi\!\left(\mathbf{h}_{[\mathrm{SEG}]}^{(k)}\right) \in \mathbb{R}^{d_m}
\label{eq:seg_proj}
\end{equation}

\noindent maps the language representation into the conditioning space of the segmentation decoder, where $d_m$ denotes the mask embedding dimension. The resulting grounding embeddings serve as semantic prompts for pixel-level localization and segmentation of agricultural entities such as diseases, pests, crops, weeds, and plant organs.

\subsection{\textbf{Grounded Segmentation Decoder}}
To enable pixel-level visual grounding, AgriScope incorporates a SAM~2-driven grounded segmentation decoder that transforms language-grounded representations into high-quality segmentation masks \citep{sam2}.
Unlike conventional segmentation models that rely on predefined object categories, the proposed decoder operates in an open-vocabulary setting and generates segmentation masks directly from natural-language expressions.

Let $\mathbf{c}_k \in \mathbb{R}^{d_m}$ denote the grounding embedding corresponding to the $k$-th \texttt{[SEG]} token generated by the LLM.
Given the dense image features extracted by the SAM~2 image encoder,

\begin{equation}
\mathbf{F}_{sam} = E_{sam}(I),
\end{equation}

\noindent the grounded segmentation decoder predicts the corresponding pixel-level mask as

\begin{equation}
M_k
=
D_{seg}
\left(
\mathbf{c}_k,
\mathbf{F}_{sam}
\right),
\label{eq:sam_decoder}
\end{equation}

\noindent where $D_{seg}(\cdot)$ denotes the SAM~2 mask decoder and $M_k \in \{0,1\}^{H \times W}$ represents the predicted segmentation mask associated with the grounded concept.

\subsection{Training Objective}
AgriScope is optimized using a joint objective that combines multimodal language understanding and pixel-level visual grounding. 
Given an agricultural image $I$, a natural-language instruction $T$, the target response sequence $Y$, and the corresponding ground-truth segmentation masks $\mathcal{M}^{gt}$, the model jointly learns textual generation and mask prediction.
The language modeling objective is defined as \citep{llava2023}:

\begin{equation}
\mathcal{L}_{LM}
=
-\sum_{i=1}^{L}
\log P(y_i \mid y_{<i}, I, T),
\label{eq:lm_loss}
\end{equation}
\noindent where $L$ denotes the length of the response sequence and $y_i$ represents the $i$-th output token.

For pixel-grounded outputs, the segmentation decoder predicts a set of masks $\mathcal{M}^{pred}=\{M_1,\ldots,M_K\}$.
Following recent grounding frameworks, segmentation supervision is provided using a combination of binary cross-entropy (BCE) and Dice losses \citep{lisa}. For a predicted mask, let $m_j^{pred}$ and $m_j^{gt}$ denote the predicted probability and ground-truth label, respectively, at pixel $j$.
The BCE loss is given by:

\begin{equation}
\begin{split}
    \mathcal{L}_{BCE}=-\frac{1}{N}\sum_{j=1}^{N}
    \left[m_j^{gt}\log(m_j^{pred})\right. \\
    \left.+ (1-m_j^{gt}) \log\left(1-m_j^{pred}\right)\right],
\end{split}
\label{eq:bce_loss}
\end{equation}

\noindent where $N$ denotes the number of pixels in the mask.
To further improve mask quality and region overlap, we employ the Dice loss \citep{milletari2016vnet}:

\begin{equation}
\mathcal{L}_{Dice}=1-\frac{2\sum_{j}m_j^{pred}m_j^{gt}+\epsilon}{\sum_{j}m_j^{pred}+\sum_{j}m_j^{gt}+\epsilon},
\label{eq:dice_loss}
\end{equation}

\noindent which directly optimizes the overlap between predicted and ground-truth masks.
The segmentation objective is therefore defined as:

\begin{equation}
\mathcal{L}_{seg}
=
\lambda_{BCE}\mathcal{L}_{BCE}
+
\lambda_{Dice}\mathcal{L}_{Dice},
\label{eq:seg_loss}
\end{equation}
\noindent The overall training objective of AgriScope combines language generation and visual grounding losses,

\begin{equation}
\mathcal{L}_{total}
=
\mathcal{L}_{LM}
+
\lambda \mathcal{L}_{seg},
\label{eq:total_loss}
\end{equation}
\noindent where $\lambda$ controls the relative importance of the grounding objective. This formulation enables AgriScope to jointly learn multimodal instruction following and pixel-grounded agricultural understanding within a unified framework.

\begin{figure*}[t!]
  \centering
  \includegraphics[width=\textwidth]{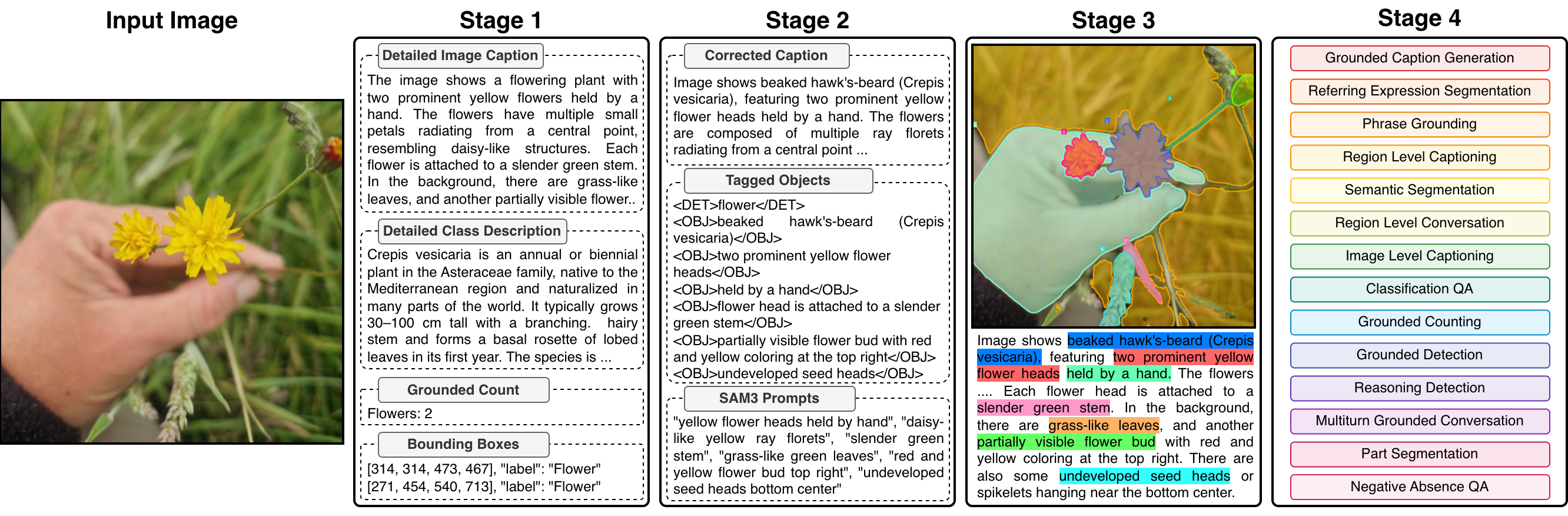}
  \caption{\textbf{Overview of the AgriGround annotation and task-generation pipeline.}
    Starting from a raw agricultural image, the pipeline first generates detailed image captions, class-level descriptions, grounded counts, and object bounding boxes. These intermediate annotations are then refined into corrected captions, explicitly tagged object mentions, and SAM~3 prompting phrases for mask generation. The resulting masks are aligned with object-level and phrase-level textual references to produce grounded captions with interleaved segmentation supervision. Finally, the verified annotations are converted into diverse instruction-following tasks, including grounded caption generation, referring expression segmentation, phrase grounding, region-level captioning, semantic segmentation, region-level conversation, image-level captioning, classification question answering, grounded counting, grounded detection, reasoning detection, multi-turn grounded conversation, part segmentation, and negative absence question answering.}
  \label{fig:AgriGround_pipeline}
\end{figure*}
 
\section{Proposed AgriGround Dataset}
\label{sec:dataset}
\subsection{\textbf{AgriGround Overview, Statistics, and Tasks}}
We introduce \textbf{AgriGround}, a large-scale pixel-grounded multimodal instruction-tuning dataset for agricultural image understanding. 
AgriGround comprises \textbf{503,919 images} and over \textbf{11.4 million} instruction-following samples collected from 22 public agricultural datasets (Table~\ref{tab:dataset_stats}) spanning plant disease diagnosis, crop and weed identification, insect pest recognition, fine-grained botanical species classification, and field-scene understanding.
To the best of our knowledge, AgriGround is the largest agricultural vision-language dataset that jointly provides natural-language descriptions, grounded object phrases, segmentation masks, and spatial annotations within a unified learning framework.
Each image is associated with four complementary supervision signals, including textual captions, grounded object phrases, pixel-level segmentation masks, and normalized bounding boxes. 
Built upon this dense supervision, AgriGround supports 14 task types: grounded caption generation (GCG), referring expression segmentation (RES), phrase grounding, region-level captioning, semantic segmentation, region-level conversation, image-level captioning, classification question answering (CQA), grounded counting, grounded detection, reasoning detection, multi-turn grounded conversation, part segmentation, and negative absence question answering. On average, each image contributes approximately 23 instruction-following samples, making AgriGround a comprehensive benchmark for grounded agricultural multimodal learning.
The complete list of tasks is illustrated in Fig.~S1 of the supplementary material.

\subsection{\textbf{AgriGround Dataset Collection and Curation}}
AgriGround is constructed by aggregating images and annotations from 22 publicly available agricultural datasets spanning plant pathology, crop and weed analysis, insect pest recognition, biodiversity monitoring, and field-scene understanding. Due to space constraints, detailed descriptions of the constituent datasets, including their names and annotation formats, are provided in the supplementary material. The source datasets provide diverse forms of supervision, including image-level labels, bounding-box annotations, and taxonomic metadata. To ensure consistency across data sources, all images and annotations are processed through a standardized curation pipeline. Images are normalized to a unified format and resolution, while heterogeneous annotations are converted into a common schema suitable for multimodal instruction tuning. Low-quality, corrupted, and excessively noisy samples are discarded to improve overall dataset quality. Table~\ref{tab:dataset_stats} summarizes the resulting statistics of AgriGround, including the train/test split, the composition of image types, and the distribution of samples across the instruction-tuning tasks derived from the source annotations.
The resulting AgriGround dataset contains \textbf{503,919 curated images} covering around \textbf{3,000 plant species}, \textbf{110 disease categories}, \textbf{102 insect pest species}, and \textbf{33 detection classes}. The curated collection forms a large-scale and diverse foundation for grounded agricultural vision-language learning, providing broad coverage across agricultural domains, imaging conditions, and biological categories.

\subsection{\textbf{AgriGround Annotation Pipeline}}
The AgriGround annotation pipeline is illustrated in Fig.~\ref{fig:AgriGround_pipeline}.
Starting from a curated collection of agricultural images and metadata, the pipeline automatically generates dense multimodal grounding annotations through four sequential stages.
\textit{First}, image captions and class-level descriptions are generated to provide rich semantic information about agricultural entities. 
\textit{Second}, the generated captions are refined and augmented with grounding tags to establish explicit associations between textual concepts and visual objects. 
\textit{Third}, grounded segmentation masks are automatically generated using the extracted concepts and spatial annotations.
\textit{Finally}, task-specific VQA pairs are synthesized, and the resulting annotations are validated through a \textit{human-in-the-loop} verification process to ensure annotation quality and consistency.

The final output of the pipeline is a large-scale pixel-grounded multimodal dataset containing aligned captions, object phrases, segmentation masks, bounding boxes, and instruction-following samples suitable for a wide range of agricultural vision-language and visual grounding tasks.

\subsubsection{\textbf{Step 1: Comprehensive Caption Generation and Class Description}}
The first stage of the annotation pipeline focuses on generating rich semantic descriptions for agricultural images and their associated classes. 
Given an input image, we employ the CapRL model \citep{caprl} to generate a detailed image-level caption that captures visual attributes, agricultural entities, spatial relationships, disease symptoms, crop structures, pest characteristics, and environmental context. 
To improve caption quality and domain relevance, CapRL is prompted using both task-specific and dataset-specific instructions, which are provided in the supplementary material.

In addition to image-level descriptions, we generate detailed class-level knowledge for all annotated categories. 
Specifically, given a ground-truth class label (e.g., plant species, disease type, insect pest, or weed category), we query the state-of-the-art Gemini~3 Pro MLLM~\citep{gemini2024} to obtain comprehensive textual descriptions, including visual characteristics, biological properties, disease symptoms, and diagnostic cues.
These class descriptions serve as an additional source of semantic supervision for downstream instruction generation and VQA tasks.
For datasets containing ground-truth bounding-box annotations, we preserve the instance-level annotations and derive object-count information directly from the labeled instances. 
These samples are subsequently utilized for grounded detection and counting tasks.
Images without bounding-box annotations are retained for other multimodal tasks, including caption generation, visual grounding, segmentation, and instruction-following learning. 
The outputs of this stage consist of image-level captions, class-level descriptions, and object-level metadata that collectively form the semantic foundation of the AgriGround annotation pipeline.

\subsubsection{\textbf{Step 2: Caption Refinement and Class Tagging}}
The second stage focuses on refining the automatically generated captions and identifying grounded agricultural concepts. 
Given the image-level captions produced by CapRL, we employ DeepSeek-V4 \citep{deepseekv4} to generate cleaner, more coherent, and domain-consistent descriptions while preserving all relevant visual information.
Caption refinement is performed using a task-specific prompt designed to improve linguistic quality and semantic completeness.

The refined captions are subsequently processed by the same language model to identify and tag salient agricultural entities, including crops, diseases, pests, weeds, plant organs, and other visual concepts. 
These entities are annotated using special grounding tokens such as \texttt{<OBJ>} and \texttt{<DET>}, establishing explicit links between textual descriptions and visual objects. 
For region-level tasks, the same procedure is applied to region-specific captions generated from selected image regions, producing grounded region descriptions and object tags. 
The outputs of this stage are refined captions and tagged object phrases that serve as the basis for subsequent visual grounding and instruction-generation stages.

\subsubsection{\textbf{Step 3: Grounded Segmentation}}
In the third stage, the tagged object and region phrases generated in the previous step are converted into textual prompts for SAM~3 \citep{sam3}.
Given a tagged concept, SAM~3 predicts the corresponding pixel-level segmentation mask, thereby establishing explicit associations between semantic concepts and image regions.
For datasets containing ground-truth bounding-box annotations, the bounding boxes are additionally used as prompts to obtain more accurate object masks. 
The output of this stage consists of grounded captions, segmentation masks, and spatial annotations that provide dense visual grounding supervision.

\subsubsection{\textbf{Step 4: Multi-Task Visual Question Answering Generation}}
The final stage generates instruction-following samples for multimodal training.
Using the image, refined captions, grounded annotations, segmentation masks, bounding boxes, and class descriptions, we automatically synthesize QA pairs covering 14 agricultural image understanding tasks.
These tasks range from image captioning and VQA to multi-turn grounded conversation, as shown in Fig.~\ref{fig:AgriGround_pipeline}.
The resulting instruction-tuning dataset contains the image reference, task specification, input prompt, target response, and corresponding spatial supervision, forming the final AgriGround dataset used for training AgriScope.

\subsection{\textbf{Human-in-the-Loop Verification}}
To ensure the quality and reliability of AgriGround, we incorporate a \textit{human-in-the-loop} verification stage following the automatic annotation pipeline. 
A team of ten domain experts with backgrounds in agriculture, plant pathology, and computer vision manually reviews a subset of the generated annotations, including image captions, grounded object phrases, segmentation masks, and instruction-following samples.

During verification, each image and its associated annotations are assessed for semantic correctness, grounding accuracy, and annotation consistency. 
Samples containing incorrect captions, inaccurate grounding, noisy masks, or ambiguous labels are either corrected or discarded.
Only high-quality verified samples are retained in the final dataset. 
This verification process significantly improves annotation reliability and reduces noise introduced during automatic annotation. 
Detailed statistics of the resulting training and testing AgriGround dataset splits are presented in Table~\ref{tab:dataset_stats}.

\subsection{Evaluation Criteria}
AgriGround is accompanied by a comprehensive evaluation protocol designed to assess multimodal understanding, visual grounding, segmentation quality, spatial localization, counting accuracy, and conversational performance. Since AgriGround supports diverse task families, we employ task-specific evaluation metrics that measure both semantic correctness and grounding accuracy.
For language generation tasks, including image captioning, grounded caption generation, and conversational understanding, we report BLEU-4, METEOR, CIDEr-D, and SPICE scores. Referring expression segmentation and semantic segmentation are evaluated using mean intersection-over-union (mIoU), Dice score, and mask recall at 0.5. For phrase grounding and grounded localization tasks, we report AP$_{50}$, mAP$_{50}$, mAP$_{75}$, and phrase-level grounding recall. Classification and question-answering tasks are evaluated using accuracy and the $F_1$ score. Finally, grounded counting performance is assessed using mean absolute error (MAE) and root mean squared error (RMSE).
Together, these metrics provide a holistic assessment of image-level, region-level, and pixel-level agricultural understanding, enabling rigorous evaluation of grounding-aware vision-language models on AgriGround.

\begin{table}[t!]
  \centering
  \small
  \caption{Statistics of the proposed AgriGround dataset.}
  \label{tab:dataset_stats}
  \scalebox{0.70}{
  \begin{tabular}{lrrrr}
    \toprule
    & \textbf{Train} & \textbf{Test} & \textbf{Total} & \textbf{Share} \\
    \midrule
    \multicolumn{5}{l}{\textbf{Overall}} \\[2pt]
    \quad Images                    & 401,234   & 102,685   & 503,919    & 79.62\% / 20.38\% \\
    \quad Samples                   & 9,095,320 & 2,325,828 & 11,421,148 & \textemdash          \\
    \quad Avg.\ samples per image   & 22.66         & 22.66         & 22.66      & \textemdash          \\
    \midrule
    \multicolumn{5}{l}{\textbf{Dataset Types (images)}} \\[2pt]
    \quad Classification            & 185,499 &  47,424 & 232,923 & 46.22\% \\
    \quad Detection                 &  22,245 &   5,693 &  27,938 &  5.54\% \\
    \quad iNatAg Subset             & 134,773 &  34,551 & 169,324 & 33.60\% \\
    \quad Insects (IP102)           &  58,717 &  15,017 &  73,734 & 14.63\% \\
    \midrule
    \multicolumn{5}{l}{\textbf{Task Distribution (samples)}} \\[2pt]
    \quad Semantic segmentation             & 1,651,229 & 422,187 & 2,073,416 & 18.15\% \\
    \quad Phrase grounding                  & 1,638,327 & 418,948 & 2,057,275 & 18.01\% \\
    \quad Referring expression segmentation & 1,638,327 & 418,948 & 2,057,275 & 18.01\% \\
    \quad Part segmentation                 &   820,240 & 209,124 & 1,029,364 &  9.01\% \\
    \quad Region-level conversation         &   749,141 & 191,757 &   940,898 &  8.24\% \\
    \quad Region-level captioning           &   722,833 & 185,023 &   907,856 &  7.95\% \\
    \quad Classification QA                 &   639,962 & 163,786 &   803,748 &  7.04\% \\
    \quad Image-level captioning            &   400,872 & 102,587 &   503,459 &  4.41\% \\
    \quad Grounded caption generation       &   379,023 &  97,007 &   476,030 &  4.17\% \\
    \quad Multi-turn grounded conversation  &   379,023 &  97,007 &   476,030 &  4.17\% \\
    \quad Grounded counting                 &    21,865 &   5,596 &    27,461 &  0.24\% \\
    \quad Grounded detection                &    21,637 &   5,535 &    27,172 &  0.24\% \\
    \quad Reasoning detection               &    21,637 &   5,535 &    27,172 &  0.24\% \\
    \quad Negative absence QA               &    11,204 &   2,788 &    13,992 &  0.12\% \\
    \bottomrule
      \end{tabular}
}
\end{table}

\tcbset{
  taskstyle/.style={
    fonttitle=\footnotesize\ttfamily\bfseries,
    colback=white,
    colframe=black!20,
    colbacktitle=black!7,
    coltitle=black,
    boxrule=0.4pt,
    arc=3pt,
    left=5pt, right=5pt, top=3pt, bottom=3pt,
    toptitle=2pt, bottomtitle=2pt,
    fontupper=\footnotesize,
  }
}

\begin{table*}[t]
\centering
\scriptsize
\caption{\textbf{Comparison of AgriScope with state-of-the-art MLLMs on seven representative AgriGround tasks.} $^\dagger$ denotes models fine-tuned on AgriGround. \textbf{Bold} and \underline{underlined} values indicate the best and second-best performance, respectively. C, ASF, Acc., $\mathcal{J}\&\mathcal{F}$, and M denote CIDEr, Agricultural Semantic Fidelity, accuracy, the average of region similarity ($\mathcal{J}$) and contour accuracy ($\mathcal{F}$), and METEOR, respectively. ``--'' indicates unsupported tasks.}
\resizebox{\textwidth}{!}{%
\setlength{\tabcolsep}{2.0pt}
\begin{tabular}{l c|cc|cc|cc|c|cc|cc|ccccc}
\toprule
\textbf{Model} & \textbf{LLM Size}
& \multicolumn{2}{c|}{\textbf{Image Cap.}}
& \multicolumn{2}{c|}{\textbf{Region Cap.}}
& \multicolumn{2}{c|}{\textbf{CQA}}
& \multicolumn{1}{c|}{\textbf{Counting}}
& \multicolumn{2}{c|}{\textbf{Sem. Seg.}}
& \multicolumn{2}{c|}{\textbf{RES}}
& \multicolumn{5}{c}{\textbf{GCG}} \\
\cmidrule(lr){3-4} \cmidrule(lr){5-6} \cmidrule(lr){7-8} \cmidrule(lr){9-9} \cmidrule(lr){10-11} \cmidrule(lr){12-13} \cmidrule(lr){14-18}
&
& C & ASF
& C & ASF
& Acc. & $F_1$
& Acc.
& mIoU & Dice
& $\mathcal{J}\&\mathcal{F}$ & cIoU
& M & C & AP$_{50}$ & mIoU & Recall \\
\midrule
\rowcolor{blue!15}
\multicolumn{18}{c}{\textbf{General MLLMs}} \\
\midrule

LLaVA-OneVision~\citep{llavaonevision}
& 7B
& 116.4 & 74.1
& 103.8 & 72.6
& 71.2 & 69.4
& 63.6
& -- & --
& -- & --
& -- & -- & -- & -- & -- \\

Qwen2.5-VL~\citep{qwen25vl}
& 7B
& 126.8 & 78.5
& 116.9 & 78.0
& 75.6 & 73.8
& 68.6
& -- & --
& -- & --
& -- & -- & -- & -- & -- \\

InternVL3~\citep{internvl3}
& 8B
& 133.2 & 80.1
& 121.4 & 79.0
& 76.8 & 74.9
& 69.2
& -- & --
& -- & --
& -- & -- & -- & -- & -- \\

Qwen3-VL-Think~\citep{qwen3vl} 
& 8B 
& 136.4 & 82.4
& 123.6 & 81.5
& 79.1 & 77.8
& 70.6
& -- & --
& -- & --
& -- & -- & -- & -- & -- \\

\midrule
\rowcolor{blue!15}
\multicolumn{18}{c}{\textbf{Grounded MLLMs}} \\
\midrule

PixelLM~\citep{pixellm}
& 7B
& 69.8 & 56.2
& 76.5 & 58.8
& 54.0 & 50.2
& 45.3
& 45.9 & 58.6
& 48.50 & 57.30
& 15.6 & 58.4 & 41.8 & 38.9 & 54.2 \\

LISA~\citep{lisa}
& 7B
& 58.4 & 48.2
& 52.9 & 46.8
& 46.7 & 42.8
& 41.6
& 38.4 & 52.1
& 44.43 & 55.70
& 13.9 & 51.6 & 39.2 & 35.7 & 50.8 \\

GLaMM~\citep{glamm}
& 7B
& 78.2 & 58.5
& 95.6 & 70.4
& 55.6 & 51.4
& 48.9
& 34.6 & 48.5
& 43.92 & 42.99
& 18.7 & 72.4 & 34.8 & 32.5 & 45.6 \\

Sa2VA~\citep{sa2va}
& 8B
& 88.9 & 66.2
& 100.7 & 72.8
& 62.8 & 60.4
& 53.5
& 47.6 & 60.3
& 51.72 & 60.95
& 20.8 & 78.5 & 43.9 & 41.7 & 58.6 \\

\midrule
\rowcolor{blue!15}
\multicolumn{18}{c}{\textbf{Agriculture-Specific and AgriGround-Tuned MLLMs}} \\
\midrule

AgriChat \citep{agrichat}
& 8B
& 10.4 & 20.13
& 43.8 & 20.08
& 37.45 & 17.98
& 13.28
& -- & --
& -- & --
& -- & -- & -- & -- & -- \\

PixelLM$^\dagger$~\citep{pixellm}
& 7B
& 105.1 & 72.0
& 101.6 & 72.9
& 72.1 & 69.8
& 65.0
& 61.7 & 74.2
& 62.42 & 68.75
& 22.4 & 88.7 & 54.1 & 50.6 & 65.2 \\

LISA$^\dagger$~\citep{lisa}
& 7B
& 91.6 & 66.1
& 84.3 & 65.0
& 70.8 & 68.2
& 63.7
& 58.7 & 71.2
& 61.85 & 68.12
& 20.5 & 83.1 & 51.8 & 49.6 & 62.4 \\

GLaMM$^\dagger$~\citep{glamm}
& 7B
& 113.7 & 73.5
& 111.8 & 77.2
& 73.6 & 71.4
& 66.4
& 55.4 & 68.5
& 58.67 & 63.84
& 23.7 & 96.8 & 55.4 & 52.1 & 65.8 \\

Sa2VA$^\dagger$~\citep{sa2va}
& 8B
& 122.0 & 78.0
& 119.8 & 80.1
& 76.4 & 74.5
& 68.8
& 63.2 & 75.6
& \underline{64.85} & \underline{70.40}
& \underline{25.8} & \underline{107.9}
& \underline{58.9} & \underline{55.0} & \underline{70.1} \\

\midrule
\rowcolor{red!15}
\textbf{AgriScope}
& 0.5B
& \textbf{146.4} & \textbf{86.7}
& \textbf{132.5} & \textbf{84.8}
& \textbf{82.4} & \textbf{80.7}
& \textbf{74.8}
& \textbf{66.1} & \textbf{78.4}
& \textbf{67.30} & \textbf{72.65}
& \textbf{27.9} & \textbf{118.6}
& \textbf{63.9} & \textbf{59.4} & \textbf{74.2} \\

\bottomrule
\end{tabular}%
}
\label{table_mainresults1}
\end{table*}
 
\begin{table*}[t]
\centering
\scriptsize
\caption{\textbf{Performance comparison of AgriScope with state-of-the-art visually grounded MLLMs on additional AgriGround tasks.} $^\dagger$ denotes models fine-tuned on AgriGround. \textbf{Bold} and \underline{underlined} values indicate the best and second-best results, respectively.}

\resizebox{\textwidth}{!}{%
\setlength{\tabcolsep}{4pt}
\begin{tabular}{l c|c|c|c|c|c|c|c}
\toprule
\textbf{Model} & \textbf{LLM Size}
& \textbf{Phrase Grounding}
& \textbf{Part Seg.}
& \textbf{Region Conv.}
& \textbf{MT Grounded Conv.}
& \textbf{Grounded Det.}
& \textbf{Reasoning Det.}
& \textbf{Negative Absence QA} \\
\cmidrule(lr){3-9}
&
& AP$_{50}$
& mIoU
& CIDEr
& CIDEr
& AP$_{50}$
& AP$_{50}$
& Acc. \\

\midrule

PixelLM$^\dagger$~\citep{pixellm}
& 7B
& 61.9
& 61.0
& 93.7
& 88.1
& 55.8
& 50.9
& 77.2 \\

LISA$^\dagger$~\citep{lisa}
& 7B
& 55.8
& 59.4
& 75.4
& 70.1
& 51.7
& 47.0
& 75.8 \\

GLaMM$^\dagger$~\citep{glamm}
& 7B
& 60.7
& 56.8
& 106.5
& 99.8
& 57.6
& 53.5
& 80.4 \\

Sa2VA$^\dagger$~\citep{sa2va}
& 8B
& \underline{64.3}
& \underline{63.0}
& \underline{112.1}
& \underline{106.4}
& \underline{61.2}
& \underline{57.3}
& \underline{82.7} \\

\midrule
\rowcolor{red!15}
\textbf{AgriScope}
& 0.5B
& \textbf{68.2}
& \textbf{65.5}
& \textbf{120.6}
& \textbf{115.2}
& \textbf{64.6}
& \textbf{61.8}
& \textbf{87.4} \\

\bottomrule
\end{tabular}%
}
\label{table_results_remaining_tasks}
\end{table*}

We comprehensively evaluate our proposed AgriScope framework across a diverse set of agricultural image analysis and understanding tasks. 
We first describe the training configuration, implementation details, evaluation datasets, and state-of-the-art comparison methods.
We then present quantitative and qualitative results, followed by extensive ablation studies and computational complexity analysis. 
The objective of these experiments is to assess the effectiveness of AgriScope in multimodal understanding relative to state-of-the-art models.

\subsection{Training and Implementation Details}
AgriScope is implemented in PyTorch and trained on NVIDIA A100 GPUs using the training split of our proposed AgriGround dataset. 
The model is initialized using pretrained BioCLIP~2 \citep{bioclip2}, DINOv3 \citep{dinoV3}, Qwen2.5-0.5B-Instruct \citep{qwen2024qwen25_05b_instruct}, and SAM~2 \citep{sam2} models.
Training is performed in two stages. 
During \textbf{Stage-I}, only the visual-language projection layers, grounding modules, and segmentation decoder are optimized, while the BioCLIP~2, DINOv3, Qwen2.5-0.5B-Instruct, and SAM~2 backbones remain frozen.
The model is trained using the AdamW optimizer \citep{adamw} with an initial learning rate of $1\times10^{-4}$, weight decay of $0.01$, batch size of $128$, and cosine learning-rate scheduling \citep{cosineLR}.
During \textbf{Stage-II}, parameter-efficient fine-tuning is performed using Low-Rank Adaptation (LoRA) \citep{lora}.
Specifically, LoRA adapters with rank $r=16$ and scaling factor $\alpha=64$ are inserted into the query and value projection layers of Qwen2.5-0.5B-Instruct, while the visual encoders remain frozen.
The model is trained on the AgriGround instruction-tuning dataset using a learning rate of $2\times10^{-5}$ and a batch size of $64$. Mixed-precision training is employed to improve computational efficiency and reduce memory consumption.
Unless otherwise specified, all experiments use an input image resolution of $448\times448$ pixels. 
The grounding decoder is optimized using the joint language modeling and segmentation objective.
Training is performed for 60 epochs during Stage-I and 10 epochs during Stage-II, with model selection based on validation-set performance.
\subsection{Datasets and Tasks}
We use the AgriGround test split to evaluate AgriScope across a wide range of pixel-grounded agricultural imagery tasks, as summarized in Table~\ref{tab:dataset_stats}.
We also conduct cross-dataset experiments in which AgriScope is trained on the proposed AgriGround dataset and evaluated on publicly available datasets, including PlantVillage \citep{plantvillage}, AGMMU \citep{agmmu}, and CDDM \citep{cddm}.

\subsection{Evaluation Metrics}
We employ standard evaluation metrics appropriate for each task \citep{schmidtova2024automatic}.
Image-level and region-level captioning are evaluated using CIDEr and Agricultural Semantic Fidelity (ASF). 
Classification question answering (CQA) and count-only object counting are evaluated using accuracy, while CQA additionally reports the $F_1$ score.
Semantic segmentation is assessed using mean intersection-over-union (mIoU) and the Dice score, whereas referring expression segmentation (RES) is evaluated using $\mathcal{J}\&\mathcal{F}$, the average of region similarity ($\mathcal{J}$) and contour accuracy ($\mathcal{F}$), and class IoU (cIoU).
For grounded caption generation (GCG), phrase grounding, part segmentation, region-level conversation, multi-turn grounded conversation, grounded detection, reasoning detection, and negative absence QA, we adopt standard metrics including METEOR, CIDEr, AP$_{50}$, mIoU, recall, and accuracy, depending on the task.

\begin{table*}[t!]
\centering
\scriptsize
\caption{\textbf{Cross-dataset generalization comparison on publicly available agricultural benchmarks.} 
AgriScope is trained on AgriGround and evaluated on PlantVillage-VQA, AGMMU, and CDDM without task-specific fine-tuning.
\textbf{Bold} and \underline{underlined} values indicate the best and second-best results, respectively.}
\resizebox{\textwidth}{!}{%
\setlength{\tabcolsep}{2.0pt}
\begin{tabular}{l c|cccc|cccc|cccc|cccc}
\toprule
\textbf{Model} & \textbf{LLM Size}
& \multicolumn{4}{c|}{\textbf{PlantVillage-VQA}}
& \multicolumn{4}{c|}{\textbf{AGMMU (MCQs)}}
& \multicolumn{4}{c|}{\textbf{AGMMU (Open-Ended)}}
& \multicolumn{4}{c}{\textbf{CDDM}} \\
\cmidrule(lr){3-6} \cmidrule(lr){7-10} \cmidrule(lr){11-14} \cmidrule(lr){15-18}
&
& BLEU-4 & ROUGE-2 & METEOR & LLM-as-Judge (\%)
& BLEU-4 & ROUGE-2 & METEOR & Accuracy (\%)
& BLEU-4 & ROUGE-2 & METEOR & LLM-as-Judge (\%)
& BLEU-4 & ROUGE-2 & METEOR & LLM-as-Judge (\%) \\
\midrule
\rowcolor{blue!15}
\multicolumn{18}{c}{\textbf{MLLMs}} \\
\midrule

LLaVA-OneVision~\citep{llavaonevision}
& 7B
& 0.14 & 0.65 & 17.25 & 57.41
& 4.20 & 7.81 & 8.88 & 70.13
& 0.35 & 2.60 & 12.45 & 52.49
& 0.45 & 2.28 & 17.17 & 55.53 \\

Qwen2.5-VL~\citep{qwen25vl}
& 7B
& 0.03 & 0.22 & 3.43 & 53.21
& 3.29 & 15.36 & 31.70 & 70.94
& 0.09 & 0.76 & 6.62 & 59.93
& 0.57 & 2.52 & 18.11 & 59.51 \\

InternVL3~\citep{internvl3}
& 8B
& 0.10 & 0.58 & 15.20 & 60.50
& 4.90 & 16.20 & 33.20 & 72.60
& 0.28 & 2.10 & 12.20 & 60.80
& 0.72 & 3.10 & 20.60 & 62.10 \\

Qwen3-VL-Think~\citep{qwen3vl}
& 8B
& 0.12 & 0.64 & 16.40 & 62.30
& 5.60 & 17.80 & 35.60 & \textbf{74.20}
& 0.32 & 2.35 & 13.20 & \underline{62.70}
& 0.78 & 3.35 & 21.80 & 63.50 \\

AgriChat~\citep{agrichat}
& 7B
& \underline{2.00} & \underline{3.18} & \underline{19.52} & \underline{74.26}
& \underline{64.94} & \underline{50.98} & \underline{63.87} & 70.19
& \underline{0.43} & \underline{2.83} & \underline{13.44} & 46.93
& \underline{6.42} & \underline{17.16} & \underline{39.59} & \underline{69.94} \\

\midrule
\rowcolor{blue!15}
\multicolumn{18}{c}{\textbf{Pixel-Grounded MLLMs}} \\
\midrule

PixelLM~\citep{pixellm} 
& 7B 
& 0.08 & 0.49 & 12.90 & 55.20 
& 3.30 & 8.90 & 17.60 & 65.80 
& 0.22 & 1.80 & 10.40 & 48.50 
& 0.44 & 2.18 & 16.60 & 53.20 \\ 

LISA~\citep{lisa} 
& 7B 
& 0.06 & 0.43 & 11.80 & 53.80 
& 2.80 & 7.60 & 15.20 & 64.20 
& 0.18 & 1.55 & 9.60 & 46.80 
& 0.39 & 2.00 & 15.90 & 51.70 \\ 

GLaMM~\citep{glamm} 
& 7B 
& 0.11 & 0.56 & 14.20 & 56.40 
& 3.70 & 9.50 & 18.80 & 66.60 
& 0.25 & 2.05 & 10.90 & 49.90 
& 0.51 & 2.42 & 17.40 & 54.80 \\

Sa2VA~\citep{sa2va}
& 8B
& 0.13 & 0.62 & 15.30 & 58.20 
& 4.10 & 10.70 & 21.30 & 68.30 
& 0.28 & 2.25 & 11.70 & 51.80 
& 0.58 & 2.70 & 18.50 & 56.80 \\

\midrule
\rowcolor{red!15}
\textbf{AgriScope}
& 0.5B
& \textbf{2.35} & \textbf{3.75} & \textbf{21.30} & \textbf{78.00}
& \textbf{66.50} & \textbf{52.40} & \textbf{65.20} & \underline{73.80}
& \textbf{0.55} & \textbf{3.20} & \textbf{14.30} & \textbf{63.50}
& \textbf{7.10} & \textbf{18.80} & \textbf{41.20} & \textbf{73.20} \\

\bottomrule
\end{tabular}%
}
\label{table_results_main_7tasks}
\end{table*}

\subsection{State-of-the-Art Models for Comparison}
\label{sec:exp:sota_models_comparison}
We compare AgriScope against three categories of state-of-the-art models: (i) general-purpose MLLMs, (ii) visually grounded MLLMs, and (iii) agriculture-specific MLLMs.
The first category includes recent general-purpose MLLMs, namely LLaVA-OneVision~\citep{llavaonevision}, Qwen2.5-VL~\citep{qwen25vl}, InternVL3~\citep{internvl3}, and Qwen3-VL-Think~\citep{qwen3vl}. 
These models are pretrained on large-scale vision-language corpora and demonstrate strong image understanding, instruction following, and multimodal conversation capabilities. 
Since they do not natively support pixel-level visual grounding, they are evaluated only on compatible tasks, including image captioning, region-level captioning, classification QA, and image-level counting.
The second category consists of visually grounded MLLMs, including PixelLM~\citep{pixellm}, Sa2VA~\citep{sa2va}, LISA~\citep{lisa}, and GLaMM~\citep{glamm}. 
These models are designed to generate language-conditioned localization and segmentation outputs, making them strong baselines for grounded caption generation, RES, phrase grounding, and semantic segmentation.
Finally, to provide a fair comparison with domain-adapted methods, we fine-tune representative grounding models, including PixelLM, Sa2VA, LISA, and GLaMM, on the proposed AgriGround dataset using the same training protocol. 
In addition, we compare against recent agricultural MLLMs such as AgriChat~\citep{agrichat}.
Unless otherwise stated, all baseline results are obtained using the official implementations and publicly released pretrained checkpoints. 
Collectively, these baselines provide a comprehensive evaluation of AgriScope against general-purpose, grounding-aware, and agriculture-specific multimodal models under a unified agricultural benchmark.

\subsection{Quantitative Results}
Table~\ref{table_mainresults1} reports the quantitative comparison between AgriScope and representative state-of-the-art general-purpose, grounding-aware, and agriculture-specific MLLMs across seven representative AgriGround tasks.
The evaluation covers image-level captioning, region-level captioning, classification question answering, count-only object counting, semantic segmentation, RES, and GCG.
Overall, AgriScope consistently achieves the best performance across all evaluation tasks despite using a substantially smaller 0.5B-parameter LLM than most competing methods.
This demonstrates the effectiveness of combining biologically informed visual representations with pixel-grounded multimodal learning for agricultural image understanding.
For image-level captioning, AgriScope achieves the highest CIDEr score of \textbf{146.4} and ASF score of \textbf{86.7}, outperforming the strongest general-purpose baseline, Qwen3-VL-Think, by \textbf{10.0} CIDEr points and improving over the best AgriGround-finetuned grounding model (Sa2VA$^\dagger$) by \textbf{24.4} CIDEr points.
Similar improvements are observed for region-level captioning, where AgriScope achieves \textbf{132.5} CIDEr and \textbf{84.8} ASF, indicating superior understanding of localized agricultural regions.
On agricultural classification QA, AgriScope obtains the highest accuracy (\textbf{82.4$\%$}) and F$_1$ score (\textbf{80.7$\%$}), demonstrating improved recognition of plant species, diseases, pests, and weeds. 
Likewise, for count-only object counting, AgriScope achieves the best counting accuracy of \textbf{74.80$\%$}, suggesting stronger quantitative reasoning over agricultural scenes.
AgriScope also establishes new state-of-the-art performance on pixel-level grounding tasks.
For semantic segmentation, it achieves an mIoU of \textbf{66.10} and Dice score of \textbf{78.40}, outperforming the strongest fine-tuned baseline, Sa2VA$^\dagger$, by 2.9 and 2.8 points, respectively.
Similar improvements are observed for RES, where AgriScope achieves the highest $\mathcal{J}\&\mathcal{F}$ (\textbf{67.30}) and cIoU (\textbf{72.65}), demonstrating more accurate localization of language-referenced agricultural objects.
The largest improvements are observed for GCG, the most challenging task requiring simultaneous language generation and pixel-level grounding. 
AgriScope achieves the best METEOR (\textbf{27.9}), CIDEr (\textbf{118.6}), AP$_{50}$ (\textbf{63.9}), mIoU (\textbf{59.4}), and recall (\textbf{74.2}), consistently outperforming all competing methods.
Compared with the strongest fine-tuned grounding model, Sa2VA$^\dagger$, AgriScope improves CIDEr by 10.7 points, AP$_{50}$ by 5.0 points, mIoU by 4.4 points, and recall by 4.1 points, highlighting its superior capability to jointly generate semantically rich descriptions and accurately ground agricultural concepts.
Table~\ref{table_results_remaining_tasks} reports the performance of AgriScope on additional AgriGround tasks, including phrase grounding, part segmentation, region-level conversation, multi-turn grounded conversation, grounded detection, reasoning detection, and negative absence QA.
These tasks further evaluate the model's ability to perform fine-grained visual grounding, multimodal interaction, and robust agricultural understanding.
AgriScope consistently achieves the best performance across all evaluated tasks, despite using a significantly smaller LLM.
Compared with the strongest baseline, Sa2VA$^\dagger$, AgriScope improves phrase-grounding AP$_{50}$ from 64.3 to \textbf{68.2}, part-segmentation mIoU from 63.0 to \textbf{65.5}, grounded-detection AP$_{50}$ from 61.2 to \textbf{64.6}, and reasoning-guided-detection AP$_{50}$ from 57.3 to \textbf{61.8}.
Furthermore, AgriScope achieves the highest CIDEr scores for both region-level conversation (\textbf{120.6}) and multi-turn grounded conversation (\textbf{115.2}), while obtaining the best accuracy (\textbf{87.4$\%$}) on the Negative Absence QA task.
These results demonstrate that the advantages of AgriScope extend beyond the representative benchmark tasks, consistently improving visual grounding, multimodal interaction, and localization performance across a wide range of agricultural image understanding tasks.
Overall, the experimental results demonstrate three important observations.
\textit{First}, general-purpose MLLMs exhibit strong language generation performance but lack pixel-level grounding capabilities. 
\textit{Second}, grounding-aware MLLMs benefit substantially from fine-tuning on AgriGround, confirming the effectiveness of the proposed dataset for agricultural visual grounding. 
\textit{Finally}, AgriScope consistently achieves the best performance across all evaluated tasks, validating the effectiveness of its biologically informed dual-encoder architecture and unified pixel-grounded multimodal learning framework.

\subsection{Cross-Dataset Generalization}
To evaluate the generalization capability of AgriScope beyond the proposed AgriGround dataset, we conduct cross-dataset experiments on three publicly available agricultural benchmarks including PlantVillage-VQA \citep{plantvillage}, AGMMU \citep{agmmu}, and CDDM \citep{cddm}. 
Following a zero-shot evaluation protocol, AgriScope is trained exclusively on AgriGround and directly evaluated on these external datasets without any task-specific fine-tuning or adaptation.
The quantitative results are summarized in Table~\ref{table_results_main_7tasks}.
Overall, AgriScope demonstrates strong cross-dataset generalization, consistently outperforming both general-purpose and agricultural MLLMs across most evaluation metrics. 
In PlantVillage-VQA, AgriScope achieves the highest BLEU-4 (2.35), ROUGE-2 (3.75), METEOR (21.30), and LLM-as-Judge score (78.0$\%$), indicating superior disease-related QA and descriptive capability. 
Similar improvements are observed on the AGMMU benchmark, where AgriScope achieves the best BLEU-4, ROUGE-2, and METEOR scores while achieving competitive multiple-choice accuracy comparable to that of much larger general-purpose MLLMs.
For the more challenging open-ended AGMMU and CDDM benchmarks, AgriScope consistently achieves the highest language-generation quality across BLEU-4, ROUGE-2, METEOR, and LLM-as-Judge evaluation.
These improvements demonstrate that the proposed AgriGround instruction-tuning dataset enables AgriScope to learn transferable agricultural visual-language representations that generalize effectively to unseen datasets, crop species, disease categories, and imaging conditions.
\textit{Overall, the cross-dataset experiments indicate that AgriScope does not simply memorize AgriGround annotations but learns robust multimodal representations that transfer well across diverse agricultural benchmarks, highlighting its applicability to real-world agricultural image understanding.}

\subsection{Visual Results}
\label{sec:qualitative_results}
\begin{figure*}[!htbp]
    \centering
    \includegraphics[width=\textwidth]{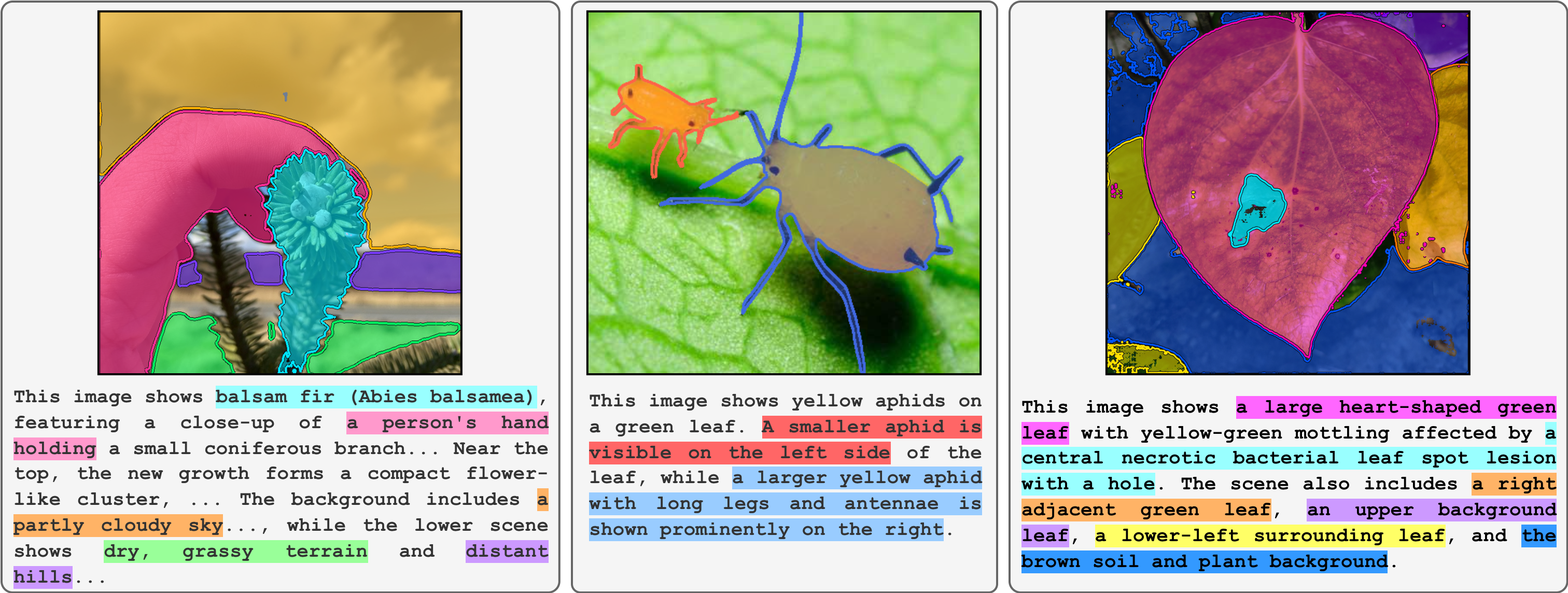}
    \caption{Qualitative grounded caption generation (GCG) results produced by AgriScope. Given the prompt ``Describe the image in detail. Please output an interleaved segmentation mask,'' the model generates detailed image descriptions while grounding each semantic concept with its corresponding pixel-level segmentation mask, demonstrating accurate localization of agricultural objects, plant structures, insect pests, disease symptoms, and scene components.}
    \label{fig:qualitative_gcg}
\end{figure*}

\begin{figure*}[!htbp]
    \centering
    \includegraphics[width=\textwidth]{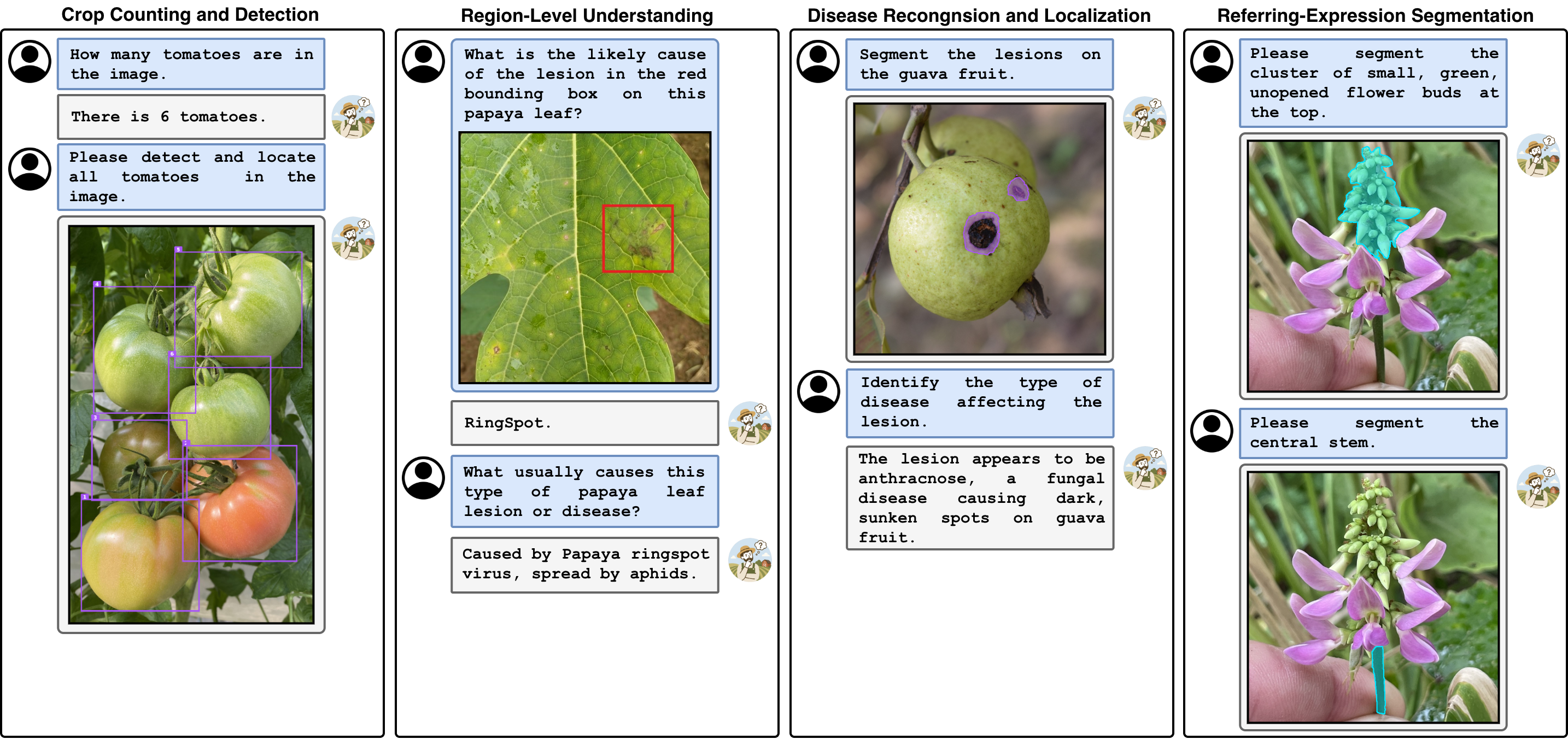}
    \caption{Qualitative results of AgriScope on representative agricultural image understanding tasks, including grounded crop counting and detection, region-level understanding, disease recognition and localization, and referring expression segmentation.}
       \label{fig:qualitative_tasks}
\end{figure*}

Figs.~\ref{fig:qualitative_gcg} and~\ref{fig:qualitative_tasks} present
representative qualitative results of AgriScope on various agricultural image
understanding tasks.
Fig.~\ref{fig:qualitative_gcg} demonstrates the GCG capability, in which AgriScope generates comprehensive image descriptions and accurately grounds semantic concepts, including plant species, insect pests, disease lesions, plant organs, and scene components, with pixel-level segmentation masks.
Fig.~\ref{fig:qualitative_tasks} showcases AgriScope on downstream tasks, including grounded crop counting and detection, region-level understanding, disease recognition and lesion localization, and referring expression segmentation.
The examples demonstrate that AgriScope can accurately interpret natural-language instructions while simultaneously producing precise localization and segmentation of agricultural objects. Overall, these results highlight the effectiveness of AgriScope in jointly performing image-level, region-level, and pixel-level agricultural understanding within a unified multimodal framework.

\subsection{Ablation Studies}

\begin{table*}[t!]
\centering
\caption{\textbf{Ablation study.}
We evaluate the contributions of the biological visual encoder, dense spatial
encoder, grounded segmentation token, agricultural instruction tuning, and
class-description enrichment.}
\scriptsize
\setlength{\tabcolsep}{4pt}
\resizebox{\textwidth}{!}{%
\begin{tabular}{lccccc}
\toprule
\textbf{Variant} & \textbf{GCG C} & \textbf{GCG mIoU} & \textbf{RES $\mathcal{J}\&\mathcal{F}$} & \textbf{RES cIoU} & \textbf{QA Overall} \\
\midrule
AgriScope without BioCLIP~2 encoder
& 109.3 & 56.7 & 64.10 & 69.60 & 76.8 \\

AgriScope without DINOv3 dense encoder
& 112.0 & 52.1 & 61.20 & 65.40 & 78.5 \\

AgriScope without \texttt{[SEG]}-conditioned mask decoding
& 114.2 & 49.0 & 58.90 & 63.10 & 79.0 \\

AgriScope without AgriGround instruction tuning
& 91.8 & 45.5 & 53.30 & 59.20 & 68.7 \\

AgriScope without class-description enrichment
& 113.6 & 57.9 & 65.50 & 70.80 & 79.1 \\

\rowcolor{red!15}
\textbf{Full AgriScope}
& \textbf{118.6} & \textbf{59.4} & \textbf{67.30}
& \textbf{72.65} & \textbf{82.4} \\

\bottomrule
\end{tabular}
}
\label{tab:ablation}
\end{table*}

Table~\ref{tab:ablation} presents an ablation analysis of the major components of AgriScope. 
We evaluate the contribution of the biological semantic encoder, dense spatial encoder, language-conditioned segmentation decoder, AgriGround instruction tuning, and class-description enrichment using Grounded Caption Generation (GCG), Referring Expression Segmentation (RES), and agricultural question answering (QA).
The results demonstrate that each component contributes positively to the overall performance. 
Removing the BioCLIP~2-based semantic encoder reduces GCG performance from 118.6 to 109.3 CIDEr and decreases QA accuracy from 82.40$\%$ to 76.8$\%$, confirming the importance of biologically informed semantic representations for agricultural understanding.
Similarly, removing the DINOv3 dense encoder causes a noticeable drop in visual grounding performance, particularly for GCG mIoU (59.4$\rightarrow$52.1) and RES ($\mathcal{J}\&\mathcal{F}$: 67.30$\rightarrow$61.20), highlighting the role of dense spatial features in precise localization.
The language-conditioned segmentation mechanism is equally important. 
Removing the \texttt{[SEG]}-conditioned mask decoding leads to the largest degradation in segmentation-oriented tasks, reducing GCG mIoU to 49.0 and RES cIoU to 63.10.
This demonstrates that explicit language-guided grounding is essential for accurate pixel-level localization.
Among all ablations, removing AgriGround instruction tuning results in the largest overall performance degradation, reducing GCG CIDEr by 26.8 points and QA accuracy by 13.7 percentage points.
This confirms that the proposed dataset provides rich multimodal supervision that substantially improves both language understanding and visual grounding. 
Finally, class-description enrichment consistently improves all evaluation tasks, indicating that incorporating detailed agricultural knowledge enhances semantic understanding and fine-grained recognition.
\textit{Overall, the ablation study demonstrates that the proposed biological semantic encoder, dense spatial encoder, language-guided segmentation decoder, and AgriGround instruction tuning are complementary components, with each contributing to the overall effectiveness of AgriScope.}

\subsection{Computational Complexity}
Table~\ref{tab:complexity} compares the computational complexity of AgriScope with representative pixel-grounded MLLMs in terms of model size, computational cost, GPU memory consumption, and inference time. 
Owing to its lightweight architecture and parameter-efficient design, AgriScope requires substantially fewer computational resources than existing grounding-aware MLLMs.
As shown in Table~\ref{tab:complexity}, AgriScope uses a \textbf{0.5B}-parameter LLM, compared with the 7B--8B language backbones used by competing methods.
Consequently, it achieves the lowest computational cost (\textbf{177 GFLOPs}) and GPU memory consumption (\textbf{6 GB}), representing approximately a \textbf{74--78\%} reduction in computation and over a \textbf{60\%} reduction in memory compared with existing grounded MLLMs.
Despite its significantly smaller architecture, AgriScope achieves an inference time of \textbf{480 ms/image}, which is comparable to the fastest competing model while consistently outperforming all baselines across agricultural multimodal understanding and visual grounding tasks. 
These results demonstrate that AgriScope provides an excellent trade-off between computational efficiency and predictive performance, making it suitable for practical precision agriculture and resource-constrained deployment.

\begin{table}[t!]
\centering
\caption{\textbf{Computational complexity comparison of AgriScope and state-of-the-art pixel-grounded MLLMs.} The comparison reports the number of model parameters, computational cost (GFLOPs), GPU memory consumption, and average inference time per image.}
\label{tab:complexity}
\setlength{\tabcolsep}{6pt}
\scalebox{0.70}{
\begin{tabular}{lcccc}
\toprule
\textbf{Model} & \textbf{LLM Params} & \textbf{GFLOPs} & \textbf{GPU Memory (GB)} & \textbf{Time/Image (ms)} \\
\midrule
PixelLM  & 7B & 685 & 15 & \textbf{480} \\
LISA     & 7B & 740 & 17 & 526 \\
GLaMM    & 7B & 772 & 17 & 2306 \\
Sa2VA    & 8B & 811 & 19 & 847 \\
\midrule
\rowcolor{red!15}
\textbf{AgriScope} & \textbf{0.5B} & \textbf{177} & \textbf{6} & \textbf{480} \\
\bottomrule
\end{tabular}
}
\end{table}

\section{Conclusion}
\label{sec:conclusion}

In this paper, we presented \textbf{AgriScope}, a unified pixel-grounded multimodal framework for agricultural image understanding. Unlike existing agricultural VLMs that primarily operate in text space, AgriScope jointly performs image-level, region-level, and pixel-level understanding, enabling a wide range of agricultural image-analysis tasks within a single architecture.
By integrating biologically informed semantic representations, dense spatial features, multimodal language understanding, and SAM~2-driven visual grounding, AgriScope establishes explicit correspondences between agricultural concepts and their visual regions, leading to more interpretable and actionable agricultural intelligence.
To support the development of AgriScope, we further introduced \textbf{AgriGround}, a large-scale pixel-grounded multimodal instruction-tuning dataset comprising 503,919 images and over 11.4 million instruction-following samples.
AgriGround provides dense supervision through aligned captions, grounded object phrases, segmentation masks, and spatial annotations across diverse agricultural domains. 
The proposed automatic annotation pipeline enables scalable generation of grounded multimodal supervision while maintaining annotation quality through \textit{human-in-the-loop} verification.
Extensive experiments demonstrate the effectiveness of AgriScope across a broad spectrum of agricultural vision-language and visual grounding tasks, consistently outperforming existing agricultural MLLMs and establishing strong baselines for future research.
We believe that AgriGround and AgriScope represent an important step toward grounding-aware agricultural artificial intelligence and will facilitate future advances in precision agriculture, crop monitoring, disease diagnosis, agricultural robotics, and multimodal decision-support systems.

\section*{Data Availability}
The datasets used in this study are publicly available, including the AgML
benchmark at \url{https://github.com/Project-AgML/AgML}. The AgriGround
annotations, AgriScope model weights, training code, and data annotation
pipeline will be made publicly available at
\url{https://github.com/boudiafA/AgriScope}.

\section*{CRediT Authorship Contribution Statement}
\textbf{Abderrahmene Boudiaf:} Conceptualization, Methodology, Investigation,
Data curation, Visualization, Writing -- original draft. \textbf{Mohamad
Alanssari:} Software, Validation, Formal analysis. \textbf{Sajid Javed:}
Supervision, Writing -- review \& editing. \textbf{Irfan Hussain:} Project
administration, Funding acquisition.

\section*{Acknowledgements}
The authors would like to thank Khalifa University of Science and
Technology for providing the computational resources and research
infrastructure that supported this work.

\section*{Declaration of Competing Interests}
The authors declare that they have no known competing financial interests or
personal relationships that could have appeared to influence the work reported
in this paper.

\bibliographystyle{elsarticle-num}
\bibliography{references}

\begin{thebibliography}{10}
\expandafter\ifx\csname url\endcsname\relax
  \def\url#1{\texttt{#1}}\fi
\expandafter\ifx\csname urlprefix\endcsname\relax\def\urlprefix{URL }\fi
\expandafter\ifx\csname href\endcsname\relax
  \def\href#1#2{#2} \def\path#1{#1}\fi

\bibitem{godfray2010}
H.~C.~J. Godfray, J.~R. Beddington, et~al., Food security: The challenge of
  feeding 9 billion people, Science 327~(5967) (2010) 812--818.
\newblock \href {https://doi.org/10.1126/science.1185383}
  {\path{doi:10.1126/science.1185383}}.

\bibitem{tilman2011}
D.~Tilman, C.~Balzer, et~al., Global food demand and the sustainable
  intensification of agriculture, Proceedings of the National Academy of
  Sciences 108~(50) (2011) 20260--20264.
\newblock \href {https://doi.org/10.1073/pnas.1116437108}
  {\path{doi:10.1073/pnas.1116437108}}.

\bibitem{agrivision}
M.~T. Chiu, et~al., {Agriculture-Vision}: {A} large aerial image database for
  agricultural pattern analysis, in: Proceedings of the CVPR, 2020, pp.
  2828--2838.

\bibitem{plantvillage}
S.~P. Mohanty, D.~P. Hughes, et~al., Using deep learning for image-based plant
  disease detection, Frontiers in Plant Science 7 (2016) 1419.
\newblock \href {https://doi.org/10.3389/fpls.2016.01419}
  {\path{doi:10.3389/fpls.2016.01419}}.

\bibitem{kamilaris2018}
A.~Kamilaris, F.~X. Prenafeta-Bol{\'{d}}u, Deep learning in agriculture: A
  survey, Computers and Electronics in Agriculture 147 (2018) 70--90.
\newblock \href {https://doi.org/10.1016/j.compag.2018.02.016}
  {\path{doi:10.1016/j.compag.2018.02.016}}.

\bibitem{ip102}
X.~Wu, C.~Zhan, et~al., {IP102}: {A} large-scale benchmark dataset for insect
  pest recognition, in: Proceedings of the CVPR, 2019, pp. 8787--8796.

\bibitem{deepweeds}
A.~Olsen, et~al., {DeepWeeds}: A multiclass weed species image dataset for deep
  learning, Scientific Reports (2019).

\bibitem{inat2021}
G.~Van~Horn, et~al., The {iNaturalist} 2021 competition dataset (2021).

\bibitem{plantdoc}
D.~Singh, N.~Jain, et~al., {PlantDoc}: A dataset for visual plant disease
  detection, in: CoDS-COMAD, 2020.

\bibitem{cropandweed}
D.~Steininger, et~al., The {CropAndWeed} dataset, in: WACV, 2023.

\bibitem{agrogpt}
M.~Awais, A.~H. S.~A. Alharthi, et~al., {AgroGPT}: Efficient agricultural
  vision-language model with expert tuning, in: Proceedings of the WACV, 2025.

\bibitem{agriclip}
M.~Nawaz, et~al., {AgriCLIP}: Adapting {CLIP} for agriculture and livestock via
  domain-specialized cross-modal pretraining, in: COLING, 2025.

\bibitem{cddm}
X.~Liu, Z.~Liu, et~al., Conversational disease diagnosis via multi-modal
  language models, in: Proceedings of the ECCV, 2024, arXiv:2503.06973.

\bibitem{agrillava}
Y.~Wang, et~al., {Agri-LLaVA}: Knowledge-infused large multimodal model for
  pest and disease diagnosis, arXiv:2412.02158 (2024).

\bibitem{agridoctor}
H.~Cheng, et~al., {AgriDoctor}: A multimodal agricultural expert system,
  arXiv:2509.17044 (2025).

\bibitem{agri3mvl}
B.~Yang, Y.~Chen, et~al., {AgriGPT-VL}: Agricultural vision-language
  understanding suite, arXiv preprint arXiv:2510.04002 (2025).

\bibitem{agrobench}
N.~I. Risa~Shinoda, et~al., {AgroBench}: Expert-annotated benchmark for
  agricultural visual question answering, in: Proceedings of the ICCV, 2025,
  arXiv:2507.20519.

\bibitem{agmmu}
H.~Gauba, et~al., {AgMMU}: A comprehensive agricultural multimodal
  understanding benchmark, in: NeurIPS Datasets and Benchmarks, 2025.

\bibitem{agromind}
H.~Li, et~al., {AgroMind}: A benchmark for agricultural remote sensing
  multimodal reasoning, arXiv:2505.12207 (2025).

\bibitem{he2016resnet}
K.~He, X.~Zhang, et~al., Deep residual learning for image recognition, in:
  Proceedings of the CVPR, 2016, pp. 770--778.

\bibitem{bioclip2}
J.~Gu, et~al., {BioCLIP~2}: Emergent properties from scaling hierarchical
  contrastive learning, in: Advances in Neural Information Processing Systems
  (NeurIPS), 2025, spotlight; arXiv:2505.23883.

\bibitem{sam3}
N.~Carion, et~al., {SAM~3}: Segment anything with concepts, arXiv preprint
  arXiv:2511.16719Under review at ICLR 2026 (2025).

\bibitem{lisa}
X.~Lai, Z.~Tian, et~al., {LISA}: Reasoning segmentation via large language
  model, in: Proceedings of the CVPR, 2024.

\bibitem{glamm}
H.~Rasheed, et~al., {GLaMM}: Pixel grounding large multimodal model, in: CVPR,
  2024.

\bibitem{osprey}
Y.~Yuan, W.~Li, et~al., {Osprey}: Pixel understanding with visual instruction
  tuning, in: Proceedings of the CVPR, 2024.

\bibitem{agrichat}
A.~Boudiaf, et~al., {AgriChat}: An agricultural multimodal large language model
  with vision-to-verified-knowledge annotation, arXiv preprint arXiv:2603.16934
  (2026).

\bibitem{redmon2016yolo}
J.~Redmon, S.~Divvala, et~al., You only look once: Unified, real-time object
  detection, in: Proceedings of the CVPR, 2016, pp. 779--788.

\bibitem{ren2015fasterrcnn}
S.~Ren, K.~He, et~al., Faster {R-CNN}: Towards real-time object detection with
  region proposal networks, in: Advances in Neural Information Processing
  Systems (NeurIPS), Vol.~28, 2015.

\bibitem{swin2021}
Z.~Liu, Y.~Lin, et~al., Swin transformer: Hierarchical vision transformer using
  shifted windows, in: Proceedings of the ICCV, 2021, pp. 10012--10022.

\bibitem{dosovitskiy2021vit}
A.~Dosovitskiy, L.~Beyer, et~al., An image is worth 16x16 words: Transformers
  for image recognition at scale, in: Proceedings of the ICLR, 2021.

\bibitem{sam}
A.~Kirillov, E.~Mintun, et~al., Segment anything, in: Proceedings of the ICCV,
  2023.

\bibitem{sam2}
N.~Ravi, V.~Gabeur, et~al., {SAM 2}: Segment anything in images and videos, in:
  Proceedings of the ICLR, 2025.

\bibitem{dinoV3}
O.~Siméoni, et~al., {DINOv3}: A scalable self-supervised vision transformer,
  arXiv preprint arXiv:2508.10104Meta AI (2025).

\bibitem{bioclip}
S.~Stevens, J.~Wu, et~al., {BioCLIP}: A vision foundation model for the tree of
  life, in: Proceedings of the CVPR, 2024, best Student Paper.

\bibitem{qwen2024qwen25_05b_instruct}
{Qwen Team}, {Qwen2.5-0.5B-Instruct},
  \url{https://huggingface.co/Qwen/Qwen2.5-0.5B-Instruct}, hugging Face model
  repository, accessed 2026-07-05 (2024).

\bibitem{qwen2025qwen25technicalreport}
{Qwen Team}, \href{https://arxiv.org/abs/2412.15115}{{Qwen2.5 Technical
  Report}} (2025).
\newblock \href {http://arxiv.org/abs/2412.15115} {\path{arXiv:2412.15115}},
  \href {https://doi.org/10.48550/arXiv.2412.15115}
  {\path{doi:10.48550/arXiv.2412.15115}}.
\newline\urlprefix\url{https://arxiv.org/abs/2412.15115}

\bibitem{llava2023}
H.~Liu, C.~Li, et~al., Visual instruction tuning, in: Advances in Neural
  Information Processing Systems (NeurIPS), Vol.~36, 2023.

\bibitem{milletari2016vnet}
F.~Milletari, N.~Navab, et~al., {V-Net}: Fully convolutional neural networks
  for volumetric medical image segmentation, in: 2016 Fourth International
  Conference on 3D Vision (3DV), 2016, pp. 565--571.
\newblock \href {https://doi.org/10.1109/3DV.2016.79}
  {\path{doi:10.1109/3DV.2016.79}}.

\bibitem{caprl}
L.~Xing, X.~Dong, et~al., {CapRL}: Stimulating dense image caption capabilities
  via reinforcement learning, arXiv preprint arXiv:2509.22647 (2025).

\bibitem{gemini2024}
{Gemini Team, Google},
  \href{https://deepmind.google/technologies/gemini/}{Gemini 3: Frontier
  multimodal intelligence}, Technical report, Google DeepMind (2025).
\newline\urlprefix\url{https://deepmind.google/technologies/gemini/}

\bibitem{deepseekv4}
{DeepSeek-AI}, {DeepSeek-V4}: Towards highly efficient million-token context
  intelligence,
  \url{https://huggingface.co/deepseek-ai/DeepSeek-V4-Pro/blob/main/DeepSeek_V4.pdf},
  technical Report. Models available at
  \url{https://huggingface.co/collections/deepseek-ai/deepseek-v4} (April
  2026).

\bibitem{llavaonevision}
B.~Li, Y.~Zhang, et~al.,
  \href{https://arxiv.org/abs/2408.03326}{Llava-onevision: Easy visual task
  transfer} (2024).
\newblock \href {http://arxiv.org/abs/2408.03326} {\path{arXiv:2408.03326}}.
\newline\urlprefix\url{https://arxiv.org/abs/2408.03326}

\bibitem{qwen25vl}
S.~Bai, K.~Chen, et~al., \href{https://arxiv.org/abs/2502.13923}{Qwen2.5-vl
  technical report} (2025).
\newblock \href {http://arxiv.org/abs/2502.13923} {\path{arXiv:2502.13923}}.
\newline\urlprefix\url{https://arxiv.org/abs/2502.13923}

\bibitem{internvl3}
J.~Zhu, W.~Wang, et~al., \href{https://arxiv.org/abs/2504.10479}{Internvl3:
  Exploring advanced training and test-time recipes for open-source multimodal
  models} (2025).
\newblock \href {http://arxiv.org/abs/2504.10479} {\path{arXiv:2504.10479}}.
\newline\urlprefix\url{https://arxiv.org/abs/2504.10479}

\bibitem{qwen3vl}
S.~Bai, Y.~Cai, et~al., \href{https://arxiv.org/abs/2511.21631}{Qwen3-vl
  technical report} (2025).
\newblock \href {http://arxiv.org/abs/2511.21631} {\path{arXiv:2511.21631}}.
\newline\urlprefix\url{https://arxiv.org/abs/2511.21631}

\bibitem{pixellm}
Z.~Ren, Z.~Huang, et~al., {PixelLM}: Pixel reasoning with large multimodal
  model, in: Proceedings of the CVPR, 2024.

\bibitem{sa2va}
H.~Yuan, X.~Li, et~al., \href{https://arxiv.org/abs/2501.04001}{Sa2va: Marrying
  sam2 with llava for dense grounded understanding of images and videos}
  (2025).
\newblock \href {http://arxiv.org/abs/2501.04001} {\path{arXiv:2501.04001}}.
\newline\urlprefix\url{https://arxiv.org/abs/2501.04001}

\bibitem{adamw}
I.~Loshchilov, F.~Hutter,
  \href{https://openreview.net/forum?id=Bkg6RiCqY7}{Decoupled weight decay
  regularization}, in: International Conference on Learning Representations
  (ICLR), 2019.
\newline\urlprefix\url{https://openreview.net/forum?id=Bkg6RiCqY7}

\bibitem{cosineLR}
I.~Loshchilov, F.~Hutter,
  \href{https://openreview.net/forum?id=Skq89Scxx}{Sgdr: Stochastic gradient
  descent with warm restarts}, in: International Conference on Learning
  Representations (ICLR), 2017.
\newline\urlprefix\url{https://openreview.net/forum?id=Skq89Scxx}

\bibitem{lora}
E.~J. Hu, Y.~Shen, et~al.,
  \href{https://openreview.net/forum?id=nZeVKeeFYf9}{Lo{RA}: Low-rank
  adaptation of large language models}, in: International Conference on
  Learning Representations, 2022.
\newline\urlprefix\url{https://openreview.net/forum?id=nZeVKeeFYf9}

\bibitem{schmidtova2024automatic}
P.~Schmidtova, S.~Mahamood, S.~Balloccu, O.~Du{\v{s}}ek, A.~Gatt, D.~Gkatzia,
  D.~M. Howcroft, O.~Pl{\'a}tek, A.~Sivaprasad, Automatic metrics in natural
  language generation: A survey of current evaluation practices, in: {NLGC},
  2024.

\end{thebibliography}

\end{document}